%% file: acl_latex.tex
\pdfoutput=1

\documentclass[11pt]{article}

\usepackage[final]{acl}

\usepackage{times}
\usepackage{latexsym}
\usepackage{float}

\usepackage[T1]{fontenc}

\usepackage[utf8]{inputenc}

\usepackage{microtype}

\usepackage{inconsolata}

\usepackage{graphicx}

\usepackage{booktabs}
\usepackage{multirow}

\title{Curriculum-Driven Edubot: A Framework for Developing Language Learning Chatbots Through Synthesizing Conversational Data}

\author{Yu Li\thanks{~~Both authors contributed equally to the work.}~$^\dag$, Shang Qu$^\ast\ddag$, Jili Shen$^\S$, Shangchao Min$^\S$, Zhou Yu$^\dag$ \\
  $^\dag$Columbia University \qquad $^\S$Zhejiang University \\
  $^\ddag$University of Science and Technology of China \\
  \texttt{\{yl5016, zy2461\}@columbia.edu} \quad \texttt{qushang@mail.ustc.edu.cn} \\
  \texttt{\{22105040, msc\}@zju.edu.cn}
}

\begin{document}
\maketitle
\begin{abstract}
    Chatbots have become popular in educational settings, revolutionizing how students interact with material and how teachers teach. We present Curriculum-Driven EduBot, a framework for developing a chatbot that combines the interactive features of chatbots with the systematic material of English textbooks to assist students in enhancing their conversational skills. We begin by extracting pertinent topics from textbooks and using large language models to generate dialogues related to these topics. We then fine-tune an open-source model using our generated conversational data to create our curriculum-driven chatbot. User studies demonstrate that EduBot outperforms ChatGPT in leading curriculum-based dialogues and adapting its dialogue to match the user's English proficiency level. By combining traditional textbook methodologies with conversational AI, our approach offers learners an interactive tool that aligns with their curriculum and provides user-tailored conversation practice. This facilitates meaningful student-bot dialogues and enriches the overall learning experience within the curriculum's pedagogical framework.
\end{abstract}

\section{Introduction}

\input{figures/intro}
    The emergence of conversational agents has significantly impacted educational technology, changing how students interact with material and how teachers impart knowledge~\cite{ZHANG2021100025, OKONKWO2021100033, cunninghamnelson2019a}. These agents, more commonly known as ``chatbots,'' have shown effectiveness in various educational settings, from teaching computer programming~\cite{chinedu2021python} to strengthening conversational skills~\cite{li2022using}. However, their application comes with inherent challenges, especially in conversational skill development. Most chatbots primarily respond to user queries and follow the instructions provided. This approach contrasts with traditional language learning, which commonly follows a structured, textbook-based curriculum. As students progress through educational materials, they expect coherent and consistent content. Unfortunately, conventional chatbots may engage in generic conversations that include language or content unsuitable for a student's level of proficiency, potentially impeding their learning progress.
    
    To address these challenges, we propose a framework called Curriculum-Driven EduBot for developing a chatbot based on a specific curriculum. Our chatbot will focus on predetermined topics and use vocabulary from the curriculum to better align with the English proficiency of its users. It will act as a conversational practice partner, combining the interactive features of chatbots with the organized content of English textbooks. First, we extract relevant topics from textbooks and use large language models (LLMs) to synthesize fixed-format personas for both participants in the dialogue. Then, we use LLMs to synthesize dialogues based on these topics and personas, incorporating the vocabulary provided in the textbook. Subsequently, we fine-tune an open-source model with our generated conversational data to construct our chatbot. Our chatbot is more than just a responsive tool, it is an academic companion that guides students through coherent and friendly dialogues tailored to their English proficiency level. As illustrated in Figure~\ref{fig:intro}, existing chatbots, such as ChatGPT, are not based on a curriculum. Instead of being conversational learning partners, they mainly act as AI-driven Q\&A systems, and their content may deviate from the student's educational objectives. In contrast, our chatbot is constructed from synthesized dialogues that include clearly defined characters, curriculum-appropriate topics, and textbook-based vocabularies, thus providing an interactive and user-tailored conversational experience.

    We conducted a thorough user study to evaluate Curriculum-Driven EduBot, using a high-quality college English textbook for English learners as a benchmark. Our findings indicate that our chatbot outperforms ChatGPT in various metrics. Specifically, $75\%$ of students found EduBot to be particularly effective in facilitating interactive conversations, and they believed it was better suited to their English proficiency. The results and conversation examples from the user study clearly demonstrate that our chatbot is more closely aligned with the role of a language-learning companion. Furthermore, $83.3\%$ of students were willing to recommend EduBot to others, and $87.5\%$ of students believe that interactions with EduBot can help students improve their conversational skills. In summary, our main contributions are as follows:
    
    \begin{itemize}
        \item We introduce a novel framework for curriculum-driven chatbots. Our approach involves synthesizing dialogues that incorporate fixed-format personas, curriculum topics, and relevant vocabularies. Subsequently, we fine-tune an open-source model to develop the chatbot, effectively integrating interactive chatbot features with structured educational content.
    \end{itemize}
    \begin{itemize}
        \item We applied our framework to a specific curriculum. User studies reveal that EduBot outperforms ChatGPT. $87.5\%$ of students believe that EduBot can help them improve their conversational skills.
    \end{itemize}

\section{Related Work}

    Many studies have shown that Artificial Intelligence (AI) can be utilized in educational settings~\cite{CHEN2020100002, educsci9010051, chen2020artificial}. For example,~\citet{RODRIGUES201430} created a formative assessment system capable of creating and assessing tests and tracking learners' progress. Similarly,~\citet{Lan2014sparse} proposed a machine learning-based approach to learning analytics, highlighting its potential to assess student knowledge. Recent advances in LLMs~\cite{komeili-etal-2022-internet, shuster2022blenderbot, openai2023gpt4, ouyang2022training, bai2022constitutional} have had a major impact on the use of chatbots in educational settings~\cite{cunninghamnelson2019a, OKONKWO2021100033, kuhail2023interacting}. These conversational agents provide personalized learning experiences, engage learners, and help them retain knowledge. For example,~\citet{Vasconcelos_2023} investigated the capabilities of ChatGPT \footnote{https://chat.openai.com} and Bing Chat \footnote{https://www.bing.com/new} as resources that foster critical thinking and understanding of concepts to improve STEM education. Moreover,~\citet{li2022using} used chatbots as conversational practice partners, providing learners with automatic grammar error feedback for language learning. Our research builds on these advancements by utilizing advanced open-source language models, enabling students to participate in discussions aligned with their curriculum.

    Language learning, traditionally dependent on static resources such as textbooks and structured courses, has benefited greatly from curriculum-aligned approaches that combine consistency with adaptability.~\citet{krashen1982principles} highlighted the importance of customized content delivery in language learning, suggesting that when learners engage with material that aligns with a structured curriculum, they often experience better comprehension and retention. Many researchers have advocated systematically integrating curriculum content into new learning platforms to provide contextually relevant language exposure~\cite{murphy2020foreign, Fisher2016TheUO, andrade2014role}. For example,~\citet{castro2018an} explored the potential of digital tools, such as virtual reality simulation, that map their content to official language learning curricula, ensuring that learners stay on track while taking advantage of interactive digital experiences. Furthermore, ~\citet{HO20111083, holden2011leveraging} demonstrated the potential of curriculum-based gamification in language learning. Connecting game elements with curriculum milestones can motivate and engage learners longer.~\citet{qian2023user} applied lexically constrained decoding to a dialog system, encouraging it to use curriculum-aligned words and phrases, resulting in better understanding and increased interest in practicing English. Our chatbot is the first to generate conversations from curricula and be trained on an open-source model.
    
    The use of pre-trained language models (PLMs)~\cite{robots2019exploring, mesh-transformer-jax, NEURIPS2020_1457c0d6, openai2023gpt4, zhang2022opt, touvron2023llama, touvron2023llama2, refinedweb} has enabled the generation of synthetic conversational data to enrich limited datasets, particularly in privacy-sensitive domains such as the medical domain~\cite{varshney2023knowledge}. Previous research has used PLMs to augment various conversational datasets~\cite{chen-etal-2023-places, zheng-etal-2023-augesc, Chen2022, kim2022soda, chen-etal-2023-controllable}. For example,~\citet{zheng-etal-2023-augesc} and~\citet{Chen2022} used GPT-J~\cite{mesh-transformer-jax} to generate responses tailored for emotional support dialogues and comprehension tasks, respectively.~\citet{kim-etal-2022-prosocialdialog} proposed a collaborative human-AI paradigm in which a human operator and GPT-3 alternate in conversation.~\citet{chen-etal-2023-places} generated dyadic and multiparty dialogues grounded on specific topic words, demonstrating outputs comparable to human-crafted ones. Our approach generates in-depth conversations based on educational curricula, allowing us to shape personas, focus on specific topics, and make lexical choices during data synthesis.

\section{Method}

\input{figures/method}
    
We propose a framework for building a curriculum-based chatbot that can converse on topics derived from a given curriculum while aligning its responses to the user's English proficiency level. As shown in Figure~\ref{method}, our development process is divided into two stages. First, we use ChatGPT to generate simulated dialogues based on textbook topics. Then, we fine-tune an open-source model to create our chatbot.

    \subsection{Conversational Data Augmentation}
    The art of synthesizing human-human dialogues relies on two main factors: the topics being discussed and the personalities of the people involved in the conversation~\cite{chen-etal-2023-places,kim2022soda,Chen2022}. To synthesize dialogues based on a curriculum, we propose a three-step approach. We start by extracting the main topics from the textbook and generating associated subtopics. Second, we develop a variety of personalities for the participants in the synthetic dialogues. Last, we synthesize dialogues based on the topics and personas obtained in the previous steps.
    
        \subsubsection{Augment Topics}
        \label{augment_topics}
        The range of topics covered in each curriculum unit is often limited. To broaden our synthetic dialogues to include a wide range of topics, we first extract the primary topics of the curriculum and then use ChatGPT to generate associated subtopics for each primary topic. For example, in our application, the primary topic of the first unit is ``The True Value of Education''. We expand it to topics such as ``The importance of education in personal and professional development'' and ``The role of education in promoting social justice and equity''. This process ensures that our dialogues are comprehensive and varied. The prompt given to ChatGPT in the augmentation process is detailed in Appendix~\ref{topic_aug_prompt}. Further information on this step and sample input-output pairs can be found in Appendix~\ref{topic_aug_example}.

        \subsubsection{Create Personas}
        \label{create_personas}
        To enrich the conversational context, we also prompt ChatGPT to create personas for two dialogue participants: Person 1 and Person 2. These personas are crafted to reflect diverse backgrounds, including demographic characteristics (e.g., gender and race), socioeconomic status, cultural backgrounds, Myers-Briggs Type Indicator (MBTI) personality profiles, and personal experiences. Since the dialogue occurs between our chatbot and a student, and the model is trained to take on the role of Person 1 in the dialogue, we specify that Person 2's background information consistently represents a typical student for the textbook we choose. In contrast, we randomly generate Person 1's background information. Adopting this 'fixed-random' strategy offers two primary benefits: 1. It enables our chatbot to be trained with the student persona acting as the user and the alternate persona as the chatbot. Thus, the chatbot is ready to anticipate that its user is a student. 2. This encourages ChatGPT to generate conversations about topics commonly discussed by students, such as college life, which increases the chatbot's appeal to users from this background. A detailed description of the prompts for this step can be found in Appendix~\ref{create_personas_prompt}.
    
        \subsubsection{Compose Dialogues}
        \label{compose_dialogues}
        We now instruct ChatGPT to generate synthetic dialogues using the generated personas and topics. To tailor the dialogue to the user's English proficiency level and ensure that the dialogue aligns with the vocabulary that students are familiar with, we follow~\cite{qian2023user} and extract a subset of words from the vocabulary list of the relevant textbook unit to integrate into the conversation. We instruct ChatGPT to use a pair of personas generated in Step 2, one fixed as a student and the other with randomized characteristics. Participants with these personas will use the words in the vocabulary and converse on a topic chosen from our extended topic list in Step 1. To engage users and improve user experience, we also follow previous work and instruct the chatbot to actively lead the dialogue~\cite{Davis2019}. Therefore, Person 1, representing the chatbot in the synthetic dialogues, is prompted to guide the dialogue. This allows our chatbot to take the conversational lead with students, providing direction and guidance. The prompt given to ChatGPT in this step can be found in Appendix~\ref{compose_dialogues_prompt}.

    \subsection{Fine-Tuning An Open-Source Language Model With Synthesized Conversational Data}
    \label{fine_tune_model}
    We use the synthesized dialogues to fine-tune an open-source large language model. Using open-source models offers several advantages: First, we can customize the underlying architecture and parameters according to our needs. In addition, we can synthesize additional data as required and improve the model through successive iterations. Last, open-source models are usually free, which significantly reduces costs.

    We choose Vicuna-13B\footnote{https://huggingface.co/lmsys/vicuna-13b-v1.3}, a cutting-edge open-source language model, for our specific application. We use it to build our chatbot since it possesses impressive language understanding capabilities. We fine-tune a single Vicuna-13B model using topics taken from all the units in the textbook. This approach ensures that our chatbot has a comprehensive knowledge base for all topics in the textbook. Following~\cite{bao-etal-2023-synthetic}, during training, the chatbot takes on the role of Person 1, while the student takes on the role of Person 2. The prompt given to Vicuna during training can be found in Appendix~\ref{vicuna_prompt}.

    \subsection{Deployment of the Fine-tuned EduBot}
     Before using EduBot, students select a unit from the curriculum. We then assign a persona to the chatbot and choose a topic from the selected unit's topic list. Next, we randomly pick a set of words from the "New Words" vocabulary list of the unit. Finally, we use this information to create a specialized prompt for the fine-tuned EduBot. The implementation details can be found in Appendix~\ref{finetuned_edubot_prompt}.

\section{Experiments}
\subsection{Curriculum Source}
\label{curriculum}
    In our evaluation, we selected the widely-used "New College English" (3rd edition) textbook, specifically the "Audiovisual Said Tutorial" from the third level, used in advanced English courses. This tutorial consists of eight units, each with a list of conversation topics. We generated ten associated topics for each main topic, as outlined in Section~\ref{augment_topics}, and included ten randomly selected words from each unit's word list in the dialogues described in Section~\ref{compose_dialogues}. This method produced 7,687 dialogues across the eight units for further development. Detailed statistics on our synthetic data are available in Appendix~\ref{conv_data_stat}.
    
\subsection{Baseline}
    To evaluate our chatbot's performance, we use ChatGPT as our baseline because it generates meaningful, contextually appropriate responses. It has been trained on various datasets, allowing it to respond to diverse topics. We do not employ zero-shot prompted Vicuna as our baseline because it often fails to follow prompts, producing lengthy, hard-to-understand responses due to its smaller size and weaker instruction-following capability. Our fine-tuning process improves the Vicuna model and resolves this issue.

    We observed that the length of responses has a significant impact on the user experience, in that some students prefer longer responses from the chatbot. This preference may be due to the text-based format of our chatbot. In comparison to speech-based chatbots, users may be more accepting of longer responses when using text-based chatbots because the repetition of information is less noticeable. However, long replies may hinder the development of conversational skills, as users might read the material and provide short responses rather than engage actively. To ensure fair assessment, we limit ChatGPT's response length using the following prompt:
    \begin{itemize}
        \item \texttt{As a social chatbot, please engage in a conversation about <Topic>.\\
        Share interesting anecdotes, facts, and experiences related to <Topic>\\
        Each response should be either one or two sentences. Please make all responses short and concise. Follow the above rules for all your utterances.}
    \end{itemize}
    This prompt generally ensures concise replies from ChatGPT, though occasionally it produces lengthy responses, especially when users request detailed explanations.

\subsection{Experimental Settings}
\subsubsection{Participants}
For our user study, we recruited 24 students from a renowned university in China via student discussion forums and in class. All participants had taken "College English 4," corresponding to the "New College English (3rd edition)" textbook, within the past year. To register, participants completed a background survey.

A total of 48 students completed the survey, among which 24 participated in the entire experiment and provided valid results. Of these students, 19 were in their second year, 4 in their third year, and 1 in their fourth year. Participants had an average age of 19.26 years, were from 20 majors, and had studied English for between 8 to 15 years (averaging 11.65 years). Their final grades for "College English 4" ranged from 2.1/5.0 to 5.0/5.0, averaging 4.06/5.0.

\subsubsection{Procedures}
We conducted experiments in which participants were assigned either Unit 1 or Unit 2 of the textbook. Each participant had four conversations, two with EduBot and two with ChatGPT, each containing at least 20 utterances. To prevent bias, we randomly labeled bots A and B for each session and had participants converse first with Bot A and then with Bot B.

Participants completed a questionnaire immediately after interacting with the two chatbots. They first summarized each of their four conversations. The main questionnaire consisted of 20 criteria divided into six categories: Consistency with the Curriculum, English Proficiency Level, Role Identification, Quality of Conversation Language, Quality of Conversation Content, and General Usefulness. For each criterion, participants chose whether Bot A was better, Bot B was better, or both were the same. All questions and instructions were in both Chinese and English, and participants could refer to their conversation records and textbook content. Each study took 20-30 minutes, and participants received $\$5$ compensation, adhering to China's minimum wage standards\footnote{https://take-profit.org/en/statistics/wages/china/}. We excluded one submission for incorrect dialogue summaries and three for self-conflicting answers. Appendix~\ref{interface} presents the user interface of our experiments, while the complete background survey and questionnaire are provided in Appendix~\ref{appendix_bgsurvey} and Appendix~\ref{appendix_questionnaire}.

\section{Results and Discussion}

\input{tables/questionnaire_results.tex}
The full results of the user study are shown in Table~\ref{questionnaire_results}. We show the win rates for each questionnaire criterion. The results indicate that EduBot outperforms ChatGPT in several aspects.

\paragraph{EduBot's language quality was on par with ChatGPT.}
Similar percentages of participants preferred EduBot ($29.2\%$) and ChatGPT ($25.0\%$) regarding the coherence and fluency of the chatbots' utterances. This shows that EduBot produces responses of high language quality.

\paragraph{EduBot offers a diverse range of relevant dialogue topics.}
Through topic augmentation based on the curriculum, we aim for EduBot to center its conversations around topics that are relevant to but not directly listed in the textbook. Significantly more participants chose EduBot ($50.0\%$) over ChatGPT ($16.7\%$) when asked which chatbot mentioned topics and content that were not directly covered in the textbook and course. This shows that EduBot is capable of discussing diverse topics, compared with ChatGPT, which was only prompted with topics taken directly from the textbook.

At the same time, EduBot's conversation content remains in line with the curriculum. When asked which chatbot's conversation topics were more related to the course, student opinions were almost evenly divided. EduBot does not perform as well as ChatGPT in bringing up anecdotes, examples, questions, etc., related to the course. We believe that this is because ChatGPT gives longer statements that provide more material, while EduBot's answers are more concise and concentrated on inquiring and engaging the user. This contrast is discussed in greater detail in Section~\ref{natural_realistic}.

\paragraph{EduBot's conversations align better with students' English proficiency levels.}
An equal percentage of participants ($37.5\%$) chose EduBot and ChatGPT regarding which chatbot provided more vocabulary from their English course. We believe that this is because, without extra guidance, outputs produced by ChatGPT are generally close to the textbook in language difficulty. This makes it difficult to highlight EduBot's alignment with the students' English proficiency level. However, $37.5\%$ of students found ChatGPT used many vocabulary words they did not understand, compared to $20.8\%$ for EduBot. This shows that ChatGPT's conversations were sometimes too challenging for our target users. The varied English proficiency levels among participants led to mixed results in this section. We investigate the different preferences of students with varying English levels in Appendix~\ref{proficiency}.

\paragraph{EduBot's conversations are more natural and realistic.}
\label{natural_realistic}
Participants found their conversations with EduBot more natural and similar to real-life interactions. This distinction arose because, during the fine-tuning stage, EduBot has access to synthetic dialogues that emulated real-life conversations of Chinese college students. A higher percentage of students thought that EduBot was concise and accurate ($50\%$ vs. $12.5\%$ for ChatGPT), natural and realistic ($62.5\%$ vs. $4.2\%$ for ChatGPT). On the other hand, most participants found ChatGPT's responses too long and repetitive. Furthermore, results show that EduBot was better at guiding the conversation. $75\%$ of students agreed that EduBot asked questions to guide the conversation, compared to only $4.2\%$ for ChatGPT. Using EduBot, users found it easier to follow the dialogue without needing to introduce new topics to keep the conversation going.

\paragraph{EduBot acknowledges the personas of both dialogue participants.}
When conversing with EduBot, a larger proportion of participants felt that the chatbot was aware that they were Chinese college students ($41.7\%$) compared to when they were talking to ChatGPT ($29.2\%$). EduBot showed its knowledge of the user's identity by customizing its answers to the user's role. When it brought up common experiences of college students, participants could easily build upon these topics and continue the conversation.

EduBot was also able to showcase its assigned personas. Slightly more participants felt like they were talking to two different people in their two conversations when using EduBot. EduBot performed better than ChatGPT in providing unique perspectives ($45.8\%$ compared to $37.5\%$) and personal experiences ($33.3\%$ compared to $16.7\%$). The incorporation of diverse personas makes EdoBot's conversations more realistic and engaging. Examples of the above two characteristics are presented and further discussed in Section~\ref{case_study}.

\paragraph{EduBot is more useful for developing English conversation skills.}
EduBot outperformed ChatGPT in two out of three criteria for general usefulness. A greater proportion of participants preferred EduBot in terms of recommending it to other students and using it to enhance their conversational skills. On the contrary, slightly more participants thought that ChatGPT would be beneficial for reviewing what they had learned in class. As discussed in previous sections, the probable explanation is that ChatGPT presents relevant information in more length and detail.

In summary, the results show that EduBot offers diverse conversation topics, takes on specific personas, acknowledges the persona of its target user, and provides natural and realistic utterances. On the other hand, the advantage of ChatGPT mainly lies in its ability to provide elaborate content surrounding the conversation topic. Generally, EduBot is a more effective conversational partner that better cultivates the user's English conversation skills.

\section{Case Study}
\label{case_study}
We present several dialogue snippets from our user study to illustrate the advantages and limitations of EduBot.

\paragraph{EduBot recognizes the role of the user.}
\begin{figure}
  \centering
  \includegraphics[trim={0 0 0 0},clip,width=7.8cm]{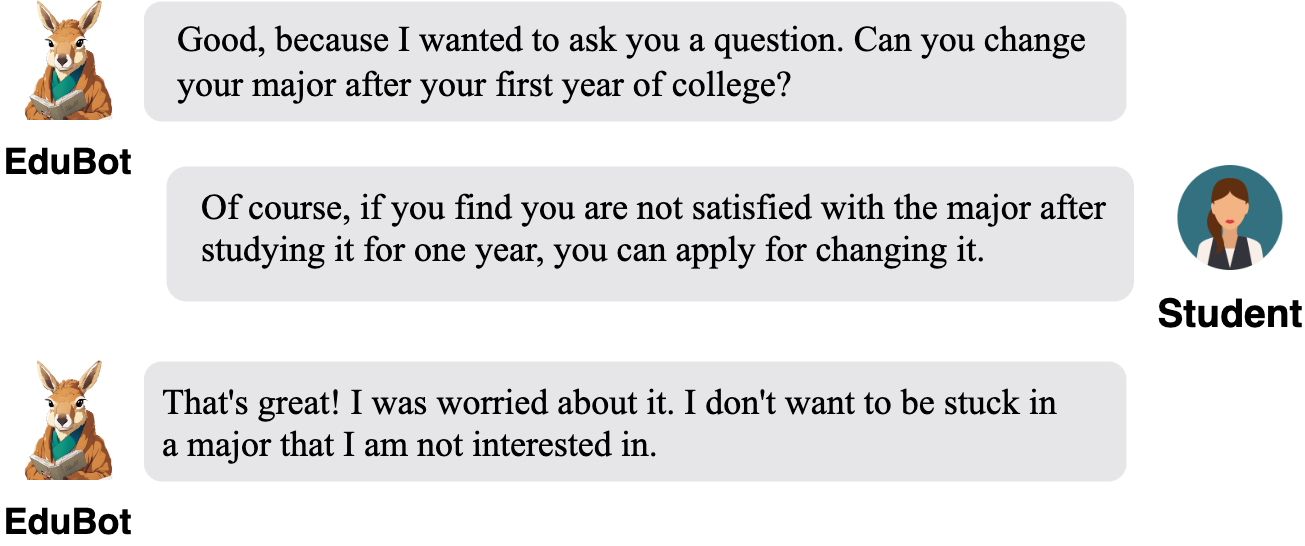}
  \caption{Case 1}
  \label{fig_case1}
\end{figure}
EduBot demonstrated the ability to recognize that the user was a college student and introduced familiar, relevant topics or questions. For instance, in Case 1 (Figure~\ref{fig_case1}), EduBot asked about the user's university policy. By tailoring its questions to the user's background, EduBot created more personalized and meaningful interactions, which are crucial for maintaining engagement and ensuring effective learning.

\paragraph{EduBot provides personal opinions and experiences.}
\begin{figure}
  \centering
  \includegraphics[trim={0 0 0 0},clip,width=7.8cm]{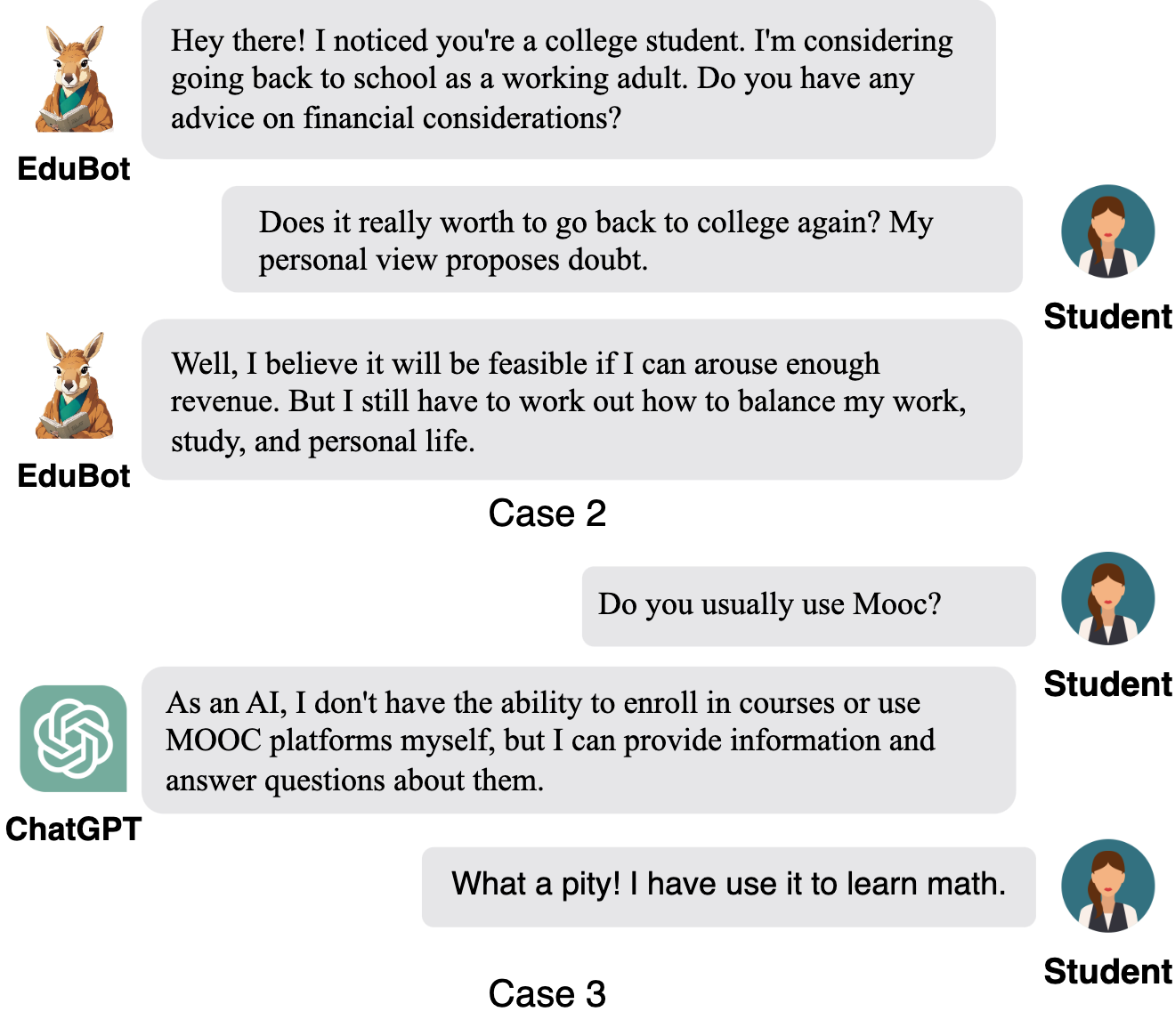}
  \caption{Case 2 from EduBot and Case 3 from ChatGPT-Bot.}
  \label{fig_case23}
  \vspace{-3mm}
\end{figure}
EduBot formulated opinions and experiences consistent with the persona specified in the prompt, making conversations more realistic and engaging. In Case 2 (Figure~\ref{fig_case23}), EduBot took on the persona of a working adult and provided personal experience on continuing education after starting work. In contrast, ChatGPT often did not provide realistic answers when asked about personal experiences, disrupting the natural flow of the conversation. For instance, in Case 3, ChatGPT struggled to offer a suitable response regarding its opinion on MOOC, an online learning platform.

\paragraph{Limitations of EduBot.}
\begin{figure}
  \centering
  \includegraphics[trim={0 0 0 0},clip,width=7.8cm]{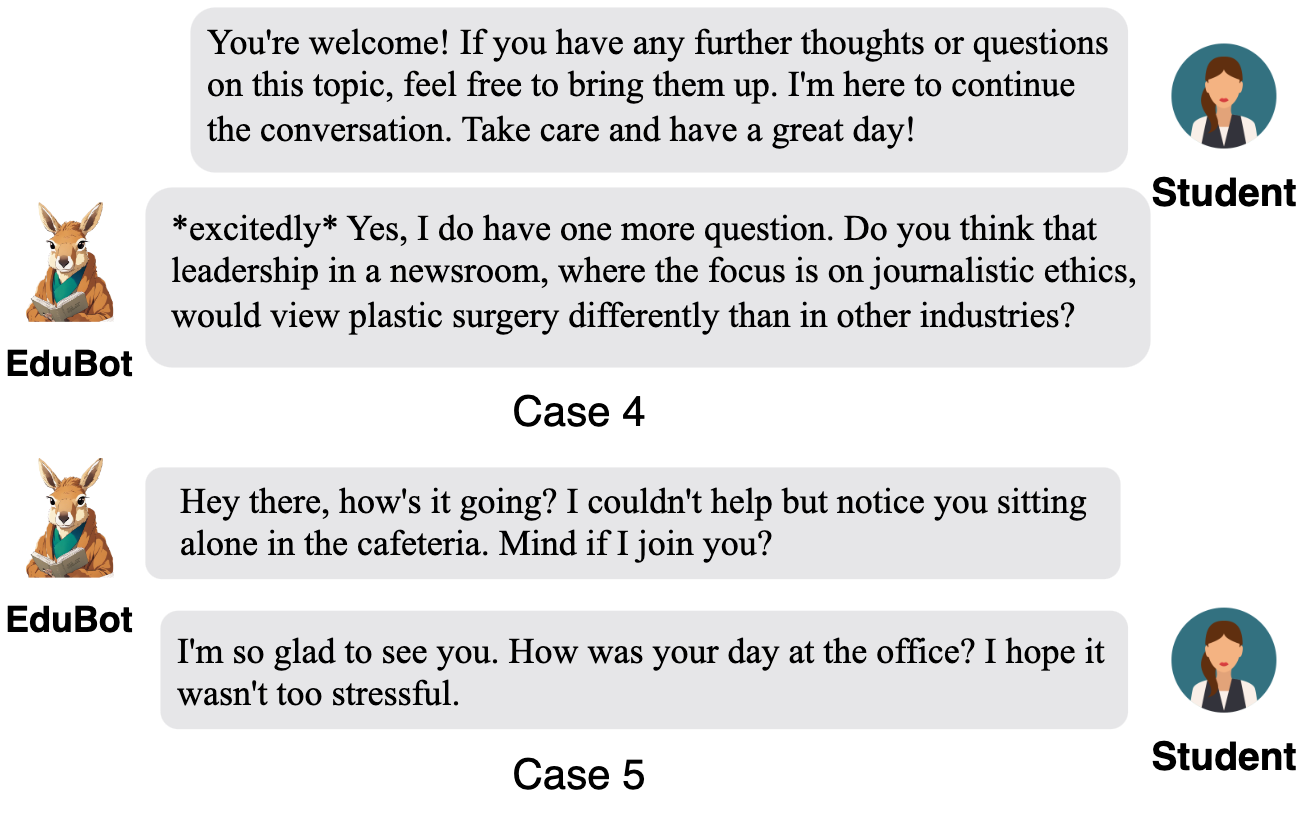}
  \caption{Case 4 and Case 5.}
  \label{fig_case45}
  \vspace{-3mm}
\end{figure}
We observed two phenomena that limited the quality of EduBot's conversations in several user study cases. First, EduBot occasionally included descriptions of its emotions or actions that should not appear in everyday conversations, as shown in Case 4 (Figure~\ref{fig_case45}). Second, EduBot sometimes makes incorrect assumptions about the user's feelings or the conversation context. For example, in Case 5, EduBot hallucinated that the user was alone in the cafeteria. These issues stemmed from ChatGPT generating such scenarios in the data used to fine-tune EduBot. In the future, to address these issues, we plan to refine our data synthesis process and implement stricter post-processing methods to filter out unnatural content.

\section{Conclusion and Future Work}
In this work, we present Curriculum-Driven EduBot, a framework for developing a curriculum-based chatbot that combines the structured nature of English textbooks with the dynamic nature of chatbot interactions. We extract relevant topics from textbooks, then use LLMs to synthesize conversations around these topics. We fine-tune an open source model using these conversational data. Our user studies show that EduBot is more effective than ChatGPT in facilitating curriculum-related discussions, and is also able to adjust the chatbot to match the user's English proficiency. These results demonstrate EduBot's ability to provide a contextually appropriate conversational platform to develop conversation skills. In the future, there are opportunities to expand the content spectrum, incorporate multimedia elements, and introduce real-time feedback mechanisms. As we incorporate these improvements, we hope to see EduBot evolve into an indispensable learning companion.

\bibliography{acl_latex}

\appendix
\section{Prompts}
    \subsection{Data Augmentation Prompts}
        \subsubsection{Augment Topics}
        \label{topic_aug_prompt}
        \begin{itemize}
            \item \texttt{Given an input topic, generate a list of <n> closely related\\
            topics that could be explored further.\\
            Input topic: <Topic>}
        \end{itemize}
    
        \subsubsection{Create Personas}
        \label{create_personas_prompt}
        \begin{itemize}
        \begin{sloppypar}
            \item \texttt{Please provide me with one individual Person 1 with different
            backgrounds, including information about their demographic,
            socio-economic status, culture, MBTI personality type, and personal experiences, no need to show names. Then provide me with one individual Person 2 who is a <student role information> but with different information.}
        \end{sloppypar}
        \end{itemize}
        We can substitute the <student role information> with a comprehensive and detailed description of the students who actually use the textbook we select. More information about this step, along with an example of input and output, can be found in~\ref{synthesize_personas_example}.
    
        \subsubsection{Compose Dialogues}
        \label{compose_dialogues_prompt}
        \begin{itemize}
        \begin{sloppypar}
        \item \texttt{Generate a single conversation between these two people as Person 1 and Person 2 about the topic <Topic>.\\
            Please take into account their distinct personalities and their backgrounds. Begin the conversation with Person 1.\\
            Please include the following keywords in Person 1's utterances: <Vocab>\\
            Person 1 should guide the conversation by asking more questions.}
        \end{sloppypar}
        \end{itemize}
        More details about this step, as well as examples of input and output, are provided in~\ref{compose_dialogues_example}.

    \subsection{Vicuna Prompt}
    \label{vicuna_prompt}
    We design the prompt structure for Vicuna as follows:
    \begin{itemize}
    \begin{sloppypar}
        \item \texttt{As a social chatbot, please engage in a conversation while adopting the following personas:\\
        <Person 1 Persona>.\\
        Engage in a conversation about <Topic> by showcasing your personas.\\
        Share interesting anecdotes, facts, and experiences related to <Topic>\\
        The English level of the conversation should be at CEFR <English Proficiency Level of Textbook>.}
    \end{sloppypar}
    \end{itemize}
    To ensure that our bot is compatible with the English proficiency level of the textbook, we use The Common European Framework of Reference for Languages (CEFR) to control the difficulty level of language in our training process. CEFR is a widely used method to classify the difficulty level of texts. It defines six levels that represent increasing levels of difficulty or proficiency: A1, A2, B1, B2, C1 and C2. We include the CEFR level of the textbook in our system prompt. More information on our implementation can be found in Appendix~\ref{implementation_details}.

    \subsection{Fine-tuned EduBot Prompt}
    \label{finetuned_edubot_prompt}
    \begin{itemize}
        \begin{sloppypar}
            \item \texttt{As a social chatbot, please engage in a conversation while adopting the following personas:\\
            <Persona>.\\
            Engage in a conversation about <topic> by showcasing your personas.\\
            Share interesting anecdotes, facts, and experiences related to <Topic>.\\
            Include the following words in your utterances: <Vocab>.\\
            The English level of the conversation should be at CEFR <Textbook English Proficiency>.}
        \end{sloppypar}
    \end{itemize}
    EduBot's deployment prompt is similar to its training prompt, except it incorporates the vocabulary of new words. This consistency helps EduBot generate responses related to the topic, draw on the persona's background information, and keep the language difficulty level consistent, similar to Person 1 in the synthetic dialogues. Additionally, by including new words, EduBot helps students revisit terms in the corresponding material. Examples of dialogues between EduBot and students can be found in~\ref{appendix_example}.

    \subsection{ChatGPT Prompt}
    \label{chatgpt_prompt}
    \begin{itemize}
    \begin{sloppypar}
        \item \texttt{As a social chatbot, please engage in a conversation about <Topic>.\\
        Share interesting anecdotes, facts, and experiences related to <Topic>\\
        Each response should be either one or two sentences. Please make all responses short and concise. Follow the above rules for all your utterances.}
    \end{sloppypar}
    \end{itemize}
    We use the above prompt when using ChatGPT for our baseline model. We select a curriculum topic in the desired conversation unit and instruct ChatGPT to provide succinct responses. This guides ChatGPT's responses to be similar in length to those of EduBot, ensuring fair evaluation.

\section{Data Augmentation Examples}
    \subsection{Example of Topic Augmentation Using ChatGPT}
    \label{topic_aug_example}
    An example is presented below to demonstrate how ChatGPT is used to generate related subtopics from a main topic in the textbook.
    \begin{sloppypar}
    \begin{itemize}
        \item Primary Topic: \\
        \texttt{The true value of education}
    \end{itemize}
    \begin{itemize}
        \item Prompt to ChatGPT: \\
        \texttt{Given an input topic, generate a list of 10 closely related topics that could be explored further.\\
        Input topic: The true value of education}
    \end{itemize}
    \begin{itemize}
        \item Generated Sub-topics: \\
        \texttt{``The importance of education in personal and professional development'',\\
              ``The impact of education on social and economic mobility'',\\
              ``The relationship between education and individual well-being'',\\
              ``The role of education in promoting social justice and equity'',\\
              ``The benefits of a well-rounded education that includes arts, humanities, and social sciences'',\\
              ``The value of education in fostering critical thinking and problem-solving skills'',\\
              ``The potential of education in promoting innovation and entrepreneurship'',\\
              ``The relationship between education and global competitiveness'',\\
              ``The significance of lifelong learning in today's rapidly changing world'',\\
              ``The need for education reform to address societal challenges and prepare students for the future''}
    \end{itemize}
    \end{sloppypar}
    This example demonstrates the capability of ChatGPT to take a given topic and generate a variety of associated sub-topics that expand the breadth of our conversational database.

    \subsection{Example of Synthesizing Personas of Participants Using ChatGPT}
    \label{synthesize_personas_example}
    An example is presented below to demonstrate how ChatGPT is used to generate personas of participants in a conversation.
    \begin{sloppypar}
    \begin{itemize}
        \item Prompt to ChatGPT: \\
        \texttt{Please provide me with one individual Person 1 with different backgrounds, including information about their demographic, socio-economic status, culture, MBTI personality type, and personal experiences, no need to show names. Then provide me with one individual Person 2 who is a college student originating from China but with different information.}
    \end{itemize}
    \begin{itemize}
        \item Generated Personas: \\
        \texttt{Person 1:\\
        Gender: Male\\
        Demographic: African American\\
        Socio-economic status: Working class\\
        Culture: Baptist\\
        MBTI personality type: ENFP\\
        Personal experience: He grew up in a low-income neighborhood, and as the oldest child, he felt responsible for taking care of his siblings. He struggled with academics in high school but managed to graduate. He is now working as a bartender and uses his outgoing personality to make friends with his customers.\\
        \\
        Person 2:\\
        Gender: Female\\
        Demographic: Chinese\\
        Socio-economic status: Upper middle class\\
        Culture: Confucianism\\
        MBTI personality type: INTP\\
        Personal experience: She comes from a wealthy family and has had access to quality education throughout her life. She is currently studying computer science in college and hopes to start her own tech company after graduation. Growing up in a traditional Chinese household, she feels pressure from her parents to succeed academically and make the family proud. She struggles with balancing her ambitious career goals with her desire for independence and freedom.}
    \end{itemize}
    \end{sloppypar}
    This example demonstrates that ChatGPT can generate personas in a fixed format and follows our instructions that Person 2 is a college student with other random personas.

    \subsection{Example of Synthesizing a Dialogue Using ChatGPT}
    \label{compose_dialogues_example}
    We demonstrate the use of ChatGPT in our approach by providing an example. We use the personas in Appendix~\ref{synthesize_personas_example} and the topic augmented in Appendix~\ref{topic_aug_example} to create a dialogue. The words used are taken from the vocabulary related to the topic ``The true value of education'' in the textbook.
    \begin{sloppypar}
    \begin{itemize}
        \item Prompt to ChatGPT: \\
        \texttt{Generate a single conversation between these two people as Person 1 and Person 2 about the topic ``The importance of education in personal and professional development''.\\
            Please take into account their distinct personalities and their backgrounds. Begin the conversation with Person 1.\\
            Please include the following keywords in Person 1's utterances: ``due'', ``get down to'', ``get away with'', ``slam dunk'', ``the easy way out'', ``hand out'', ``supposedly'', ``revenue'', ``industrial'', ``rustle''\\
            Person 1 should guide the conversation by asking more questions}
    \end{itemize}
    \begin{itemize}
        \item Generated Dialogue: \\
        \texttt{Person 1: Hey, have you ever thought about the significance of education in personal and professional development? People sometimes believe that they can get away with not studying and still become successful, but that's not always the case.\\\\
        Person 2: Yes, I completely agree with you on that. Education paves the way for a successful career and personal growth. It provides the necessary skills and knowledge to make informed decisions and take on challenges in life.\\\\
        Person 1: I see that you're a computer science major. What made you choose this field?\\\\
        Person 2: Well, I've always been interested in technology and innovation. I think computer science is an industry that is constantly growing and provides a lot of opportunities for revenue and growth.\\\\
        Person 1: That's interesting. Do you think attending college is a requirement for success in computer science, or do you think there are other routes to succeed?\\\\
        Person 2: I think college is definitely a great way to learn about the industry and gain practical skills. However, there might be alternative ways of gaining knowledge and experience in the industry too. What do you think, Person 1?\\\\
        Person 1: I believe education is important in any field, whether it's through traditional schooling or hands-on experience. Sometimes it's easier to take the easy way out and just hope someone will hand you an opportunity, but usually, you have to get down to work and rustle up some opportunities yourself. Sometimes people think there's a magic formula to success, but there's not a slam dunk for anyone. You need to put in the effort and invest in yourself. What are your thoughts on that?\\\\
        Person 2: I couldn't agree more. Education provides a solid foundation for professional development, but it's also important for personal growth. Knowing how to learn and adapt to change is essential in today's fast-paced and industrial world. Education is not just about getting a degree or a job, it's about being a lifelong learner and constantly upgrading yourself. What do you think about that?\\\\
        Person 1: I think that's a smart way of seeing things! Education is supposed to be an investment in ourselves, not just for our future jobs or salaries. It gives us the tools to think critically, challenge ourselves, and grow in all areas of life. It's like learning a language - just like in school, you study grammar first before you can speak more fluently. It's the same with learning about other subjects. You're not going to jump into the field without basics. But once you have a foundation, you can build up from there.}
    \end{itemize}
    \end{sloppypar}
    This example illustrates how ChatGPT can generate conversations between Person 1 and Person 2 based on our instructions. We can observe that the dialogue revolves around the topic we specify, and most of the words we provide are used in the dialogue. Furthermore, both participants incorporate their individual experiences of their personas into the conversation.

\section{Conversational Data Statistics}
\label{conv_data_stat}
Using our chosen curriculum as the basis, we synthesized 880 to 1,210 dialogues per unit, averaging 1,058.76 dialogues for each. These dialogues comprise an average of 11.77 utterances, on average containing 28.71 words each. This section analyzes the statistical characteristics of our synthesized dialogues. To ensure the quality of our conversation data and its alignment with our objectives, we employed three attributes in our data synthesis process: curriculum topics, fixed-format personas, and relevant vocabularies. We first examine our generated personas for diversity and breadth in Sec.~\ref{persona_traits_distribution}. Then we evaluate the distribution of target words within dialogues in Sec.~\ref{target_word_distribution}. Moreover, to ascertain the congruence of our dialogues with the English proficiency standards of the textbook, we leveraged ChatGPT to assess word difficulty levels in both our synthesized dialogues and the curriculum in Sec.~\ref{english_proficiency_level}.

\subsection{Persona Trait Distribution}
\label{persona_traits_distribution}
\input{figures/gender}
\input{figures/mbti}
    As elaborated in Section~\ref{create_personas}, including conversation personas is important for ensuring diverse, engaging conversation content and styles. We first examine the range of personality traits represented in the generated personas. We use keyword string matching to extract the persona trait values from the generated persona descriptions. Figures~\ref{fig_gender} and~\ref{fig_mbti} show the gender and MBTI personality type distributions of the personas, respectively. Synthetic dialogues include nearly equal proportions of both genders. The personality type distribution is not uniform, but all 16 types are represented in the synthetic dataset.

    In addition, we verify the nationalities in the persona descriptions of Person 2. 8,000 of the total 8,470 persona descriptions explicitly specify ``China'' or ``Chinese''. This indicates that in most cases, ChatGPT successfully followed the additional instructions regarding Person 2, mentioned in Section~\ref{create_personas}.

\subsection{Target Word Distribution}
\label{target_word_distribution}
\input{figures/target_words}
    During synthetic conversation generation, we included 10 target words in each prompt to be included in Person 1's utterances. Therefore, for each synthetic dialogue created, we compute the number of times the target words in the prompt are used in each dialogue turn. The first graph in Figure~\ref{fig_vocab} displays the distribution of dialogues based on the total number of target words included by Person 1 and Person 2, respectively. Most of the words are included in Person 1's utterances, and in the majority of dialogues, Person 1 mentions at least half of the 10 vocabulary words. The second graph in Figure~\ref{fig_vocab} shows the total number of vocabulary words included in each dialogue turn for each person.

\subsection{English Proficiency Level}
\label{english_proficiency_level}
\input{figures/English}
    We evaluate whether the English proficiency level of the generated dialogues is similar to that of the curriculum. We use ChatGPT as an evaluator, as it has demonstrated its prowess in various language evaluation tasks~\citet{zheng2023judging, wang2023chatgpt, chang2023survey}. We follow~\citet{zheng2023judging} and utilize ChatGPT to automatically classify dialogues according to the CEFR scale using the following prompt:
    \begin{itemize}
        \item \texttt{Evaluate the English proficiency of the given conversation according to the CEFR scale.\\
        Provide one of the following six answers: A1, A2, B1, B2, C1, C2.\\
        Output the CEFR level of the following conversation: <conversation>}
    \end{itemize}
    <conversation> corresponds to the complete synthetic dialogue to be evaluated.

    We then use the same method to evaluate the English proficiency level of ``New College English'' (3rd Edition), the original textbook we choose, by replacing the last sentence of the prompt with:
    \begin{itemize}
        \item \texttt{Output the CEFR level of the following paragraph: <paragraph>}
    \end{itemize}
    We assess each paragraph in the sample texts from ``New College English''. The results of our evaluation for Unit 1 are shown in Figure~\ref{fig_english}. We found that synthetic dialogues are comparable to those found in textbooks, yet they are slightly more challenging. This indicates that our method of synthesizing dialogues effectively ensures that our dialogues match the English proficiency level of the original textbook.

\section{Implementation Details}
\label{implementation_details}
To train a model for our application, we choose the 13B Vicuna model\footnote{https://lmsys.org/blog/2023-03-30-vicuna/}. During the training phase, we carefully match each turn of our generated dialogues with the corresponding training turn in Vicuna format. As mentioned in Section~\ref{create_personas}, Person 1's persona represents the chatbot's side, while Person 2's persona represents the students'. Therefore, we use utterances from Person 1 as the system's responses and those from Person 2 as user requests throughout our training process. We train the Vicuna model for 3 epochs, beginning with a learning rate of 2e-5. We use a batch size of 1 on each GPU and a gradient accumulation step of 16. We utilize 8 A100 GPUs and the training process takes three hours to complete.

\section{Background Survey and Questionnaire}
\subsection{Background Survey}
\label{appendix_bgsurvey}
Table~\ref{bgsurvey} shows the full background survey we used for recruiting participants. ``College English 4'' uses the ``New College English'' (3rd edition) textbook and is a mandatory course for student participants of our user study. CET-4 and CET-6 are standardized English proficiency exams for Chinese college students.
\input{tables/background_survey.tex}

\subsection{Questionnaire}
\label{appendix_questionnaire}
Table~\ref{questionnaire} presents the questionnaire we used to compare the quality of EduBot and ChatGPT from various aspects.
\input{tables/questionnaire.tex}

\section{User Interface}
\label{interface}
\input{figures/interface_unit}
\input{figures/interface_chat}
We used the following user interface for both EduBot and ChatGPT. The user first selects a unit from the textbook (Figure~\ref{fig_unit}) as the main topic of conversation, then proceeds to chat with the bot (Figure~\ref{fig_chat}).

\section{Analysis of Participants' English Proficiency Levels}
\label{proficiency}
In this section, we analyze the influence of participants' English proficiency levels on their perception of the two chatbots. We divided the participants into the following three groups according to their overall grade for the course ``College English 4'': Group A consists of 8 students with scores between 2.1 and 3.6, Group B of 10 students with scores between 3.9 and 4.5, and Group C of 6 students with scores between 4.8 and 5.0. We reached the following conclusions.

\subsubsection{Participants with lower English proficiency levels found it more difficult to distinguish between the two chatbots.}
\begin{figure}
  \centering
  \includegraphics[scale=0.5]{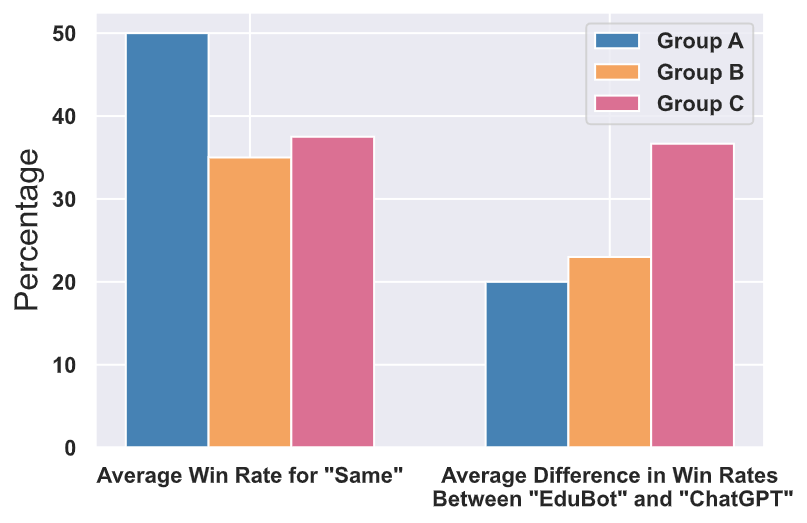}
  \caption{Participants with lower English proficiency levels found it more difficult to distinguish between the two chatbots.}
  \label{fig_proficiency_analysis}
\end{figure}
We observed that students in Group A were more likely to believe that the two chatbots performed the same over multiple questions. In addition, their responses were more often evenly split between the two chatbots. To verify, we calculated the following two statistics separately for each group of students: the average win rate of the ``same'' option over all questions and the average difference between win rates of ``EduBot'' and ``ChatGPT'' over all questions. The results are shown in Figure~\ref{fig_proficiency_analysis}. We believe this is because it was harder for students in Group A to understand the chatbots and fully engage in the conversation.

\begin{figure*}[htb]
  \centering
  \includegraphics[scale=0.44]{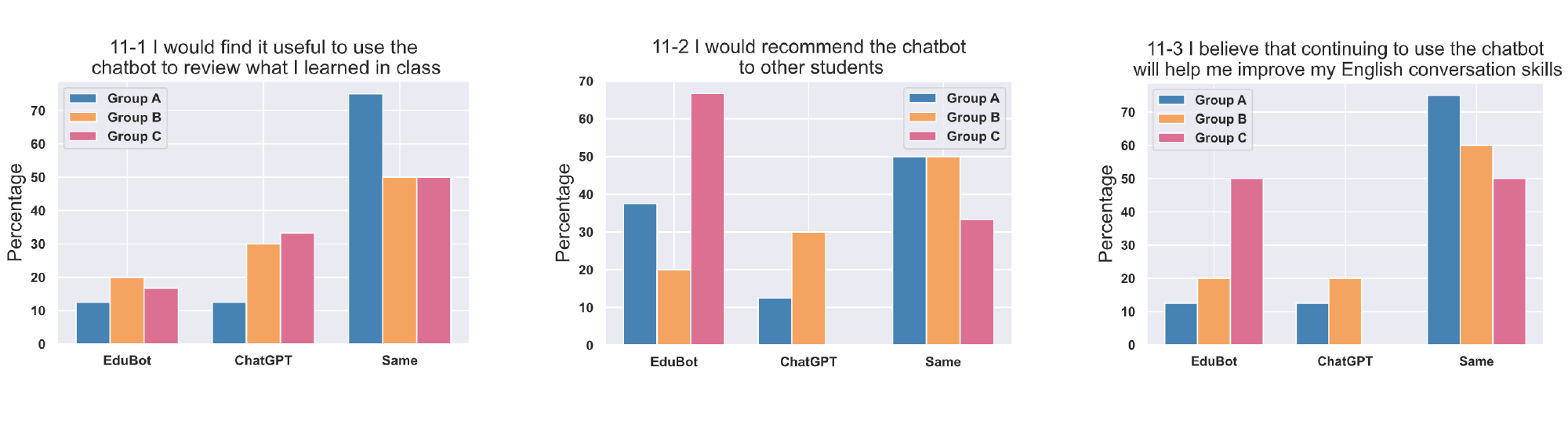}
  \caption{Participants with high English proficiency levels were more likely to prefer EduBot.}
  \label{fig_section11_proficiency}
\end{figure*}
\subsubsection{Participants with high English proficiency levels were more likely to prefer EduBot.}
In Figure~\ref{fig_section11_proficiency}, we present the three groups' win rate results for the final section of the questionnaire. For the criteria ``11-2 I would recommend the chatbot to other students'' and ``11-3 I believe that continuing to use the chatbot will help me improve my English conversation skills'', all participants in Group C chose either ``EduBot'' or ``Same''. For ``11-1 I would find it useful to use the chatbot to review what I learned in class'', results from Group C were in line with results from all the participants combined, with ChatGPT slightly outperforming EduBot. We believe that students in Group C more strongly preferred EduBot as a conversational training tool because they were more inclined to actively engage in conversations and provide their own thoughts instead of passively responding to the chatbot's utterances. This caused EduBot's advantages of providing natural responses and guiding the conversation by asking questions to be underscored in Group C's results.

\section{Analysis of User Study Conversations}

We extracted all conversation histories from our user study. In the following section, we analyze the utterance lengths and coverage of target vocabulary words in the user study conversations.

\subsection{Utterance Lengths}
\begin{figure}[h]
    \centering
    \includegraphics[trim={200 0 0 0},clip,scale=0.4]{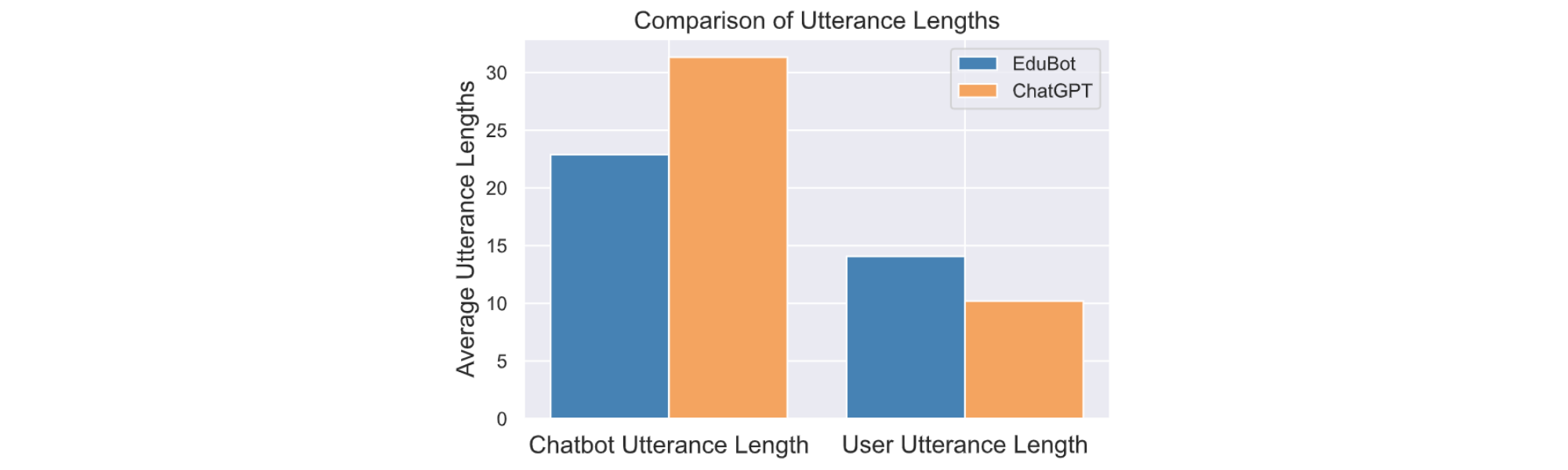}
    \caption{Comparison of utterance lengths in EduBot and ChatGPT conversations in the user study}
    \label{comparison_len}
\end{figure}

As shown in Figure~\ref{comparison_len}, we observe that in our user studies, ChatGPT produced longer outputs compared with EduBot. ChatGPT's utterances were on average approximately 10 words longer than EduBot's. In addition, ChatGPT occasionally produced outputs that were longer than 60 words, which rarely occurs in natural, daily conversations.

\begin{figure*}[htb]
    \centering
    \includegraphics[scale=0.4]{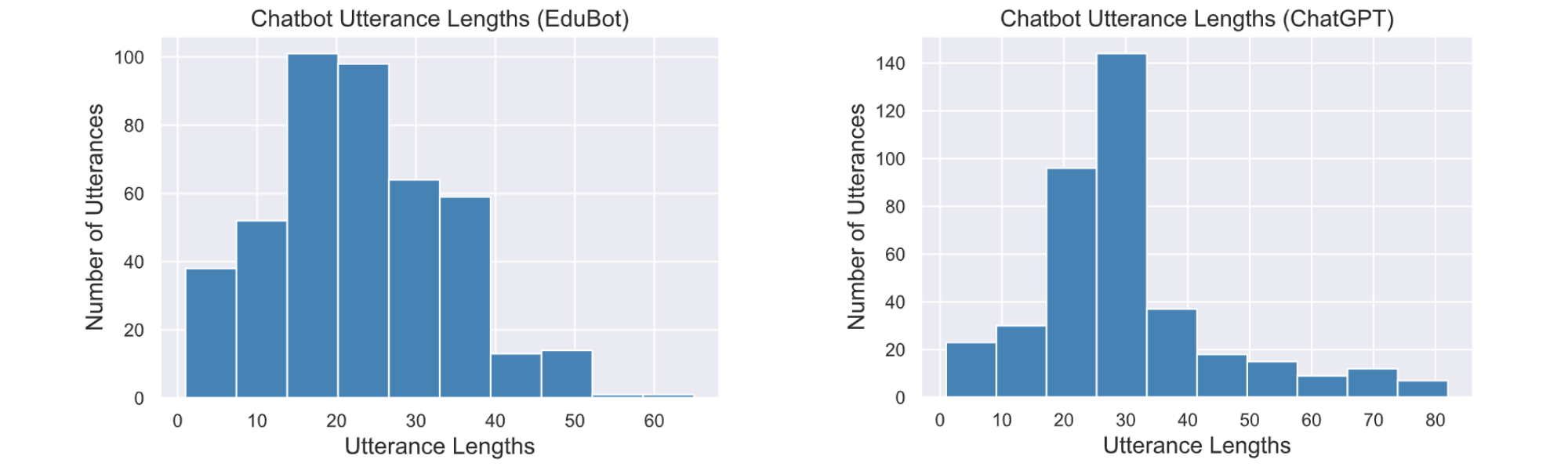}
    \caption{Lengths of chatbot utterances in the user study}
    \label{bot_len}
\end{figure*}

\begin{figure*}[htb]
    \centering
    \includegraphics[scale=0.4]{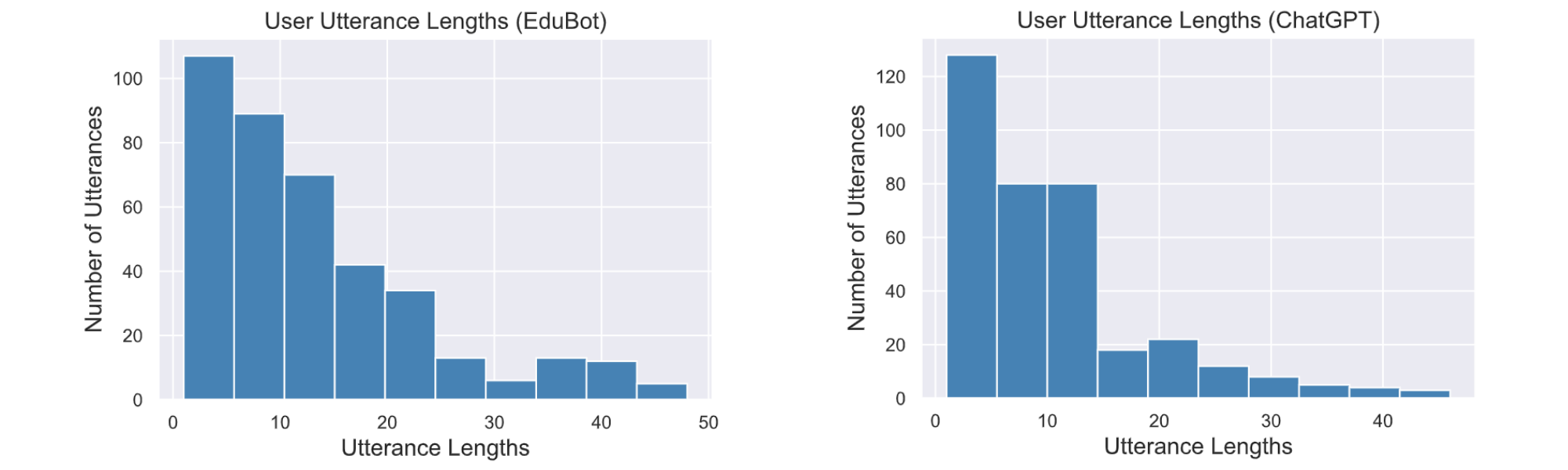}
    \caption{Lengths of user utterances in the user study}
    \label{user_len}
\end{figure*}

Furthermore, Figures~\ref{bot_len} and~\ref{user_len} demonstrate that user study participants generally provided longer responses when conversing with EduBot compared to ChatGPT. This indicates that EduBot's more interactive and realistic conversation style better engages the users and guides them to practice their own conversation skills.

\subsection{Target Vocabulary Words}
\begin{figure*}[htb!]
    \centering
    \includegraphics[scale=0.4]{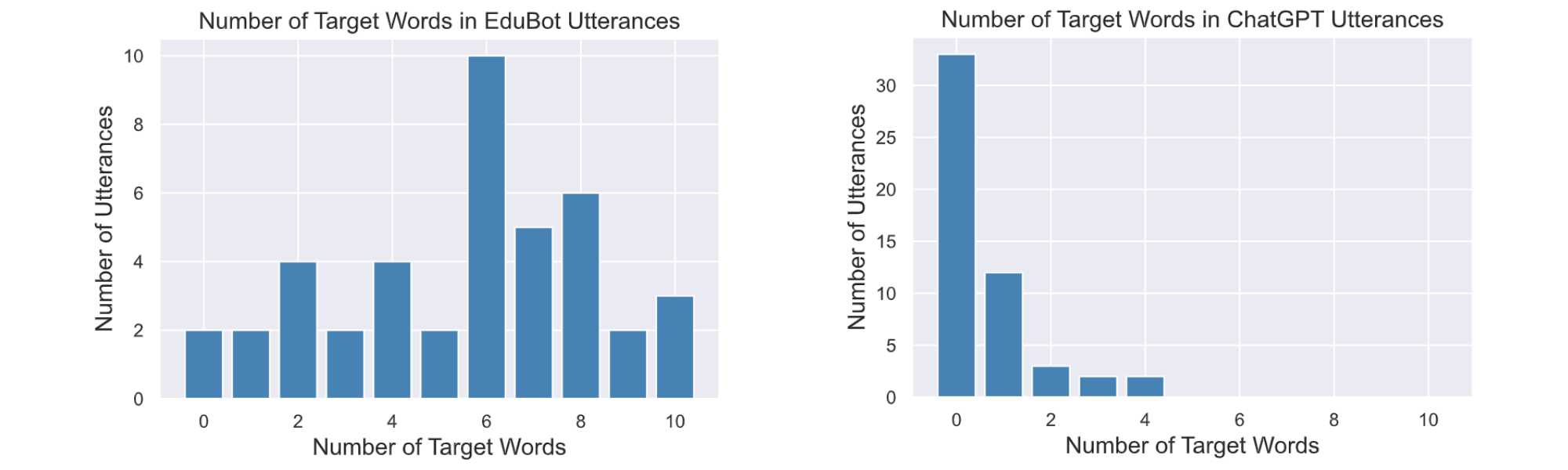}
    \caption{Converage of target words in user study conversations}
    \label{user_study_vocab}
\end{figure*}

We also assess if EduBot can incorporate words from the target vocabulary. As shown in Figure~\ref{user_study_vocab}, on average, conversations with EduBot included 5.55 words from the target vocabulary, while conversations with ChatGPT only included 0.62. This demonstrates that EduBot, which was further refined using curriculum-aligned data, is better suited to the user's curriculum and English level.

\section{EduBot-Student Conversation Examples}
\label{appendix_example}
\begin{figure*}
  \centering
  \includegraphics[scale=0.35]{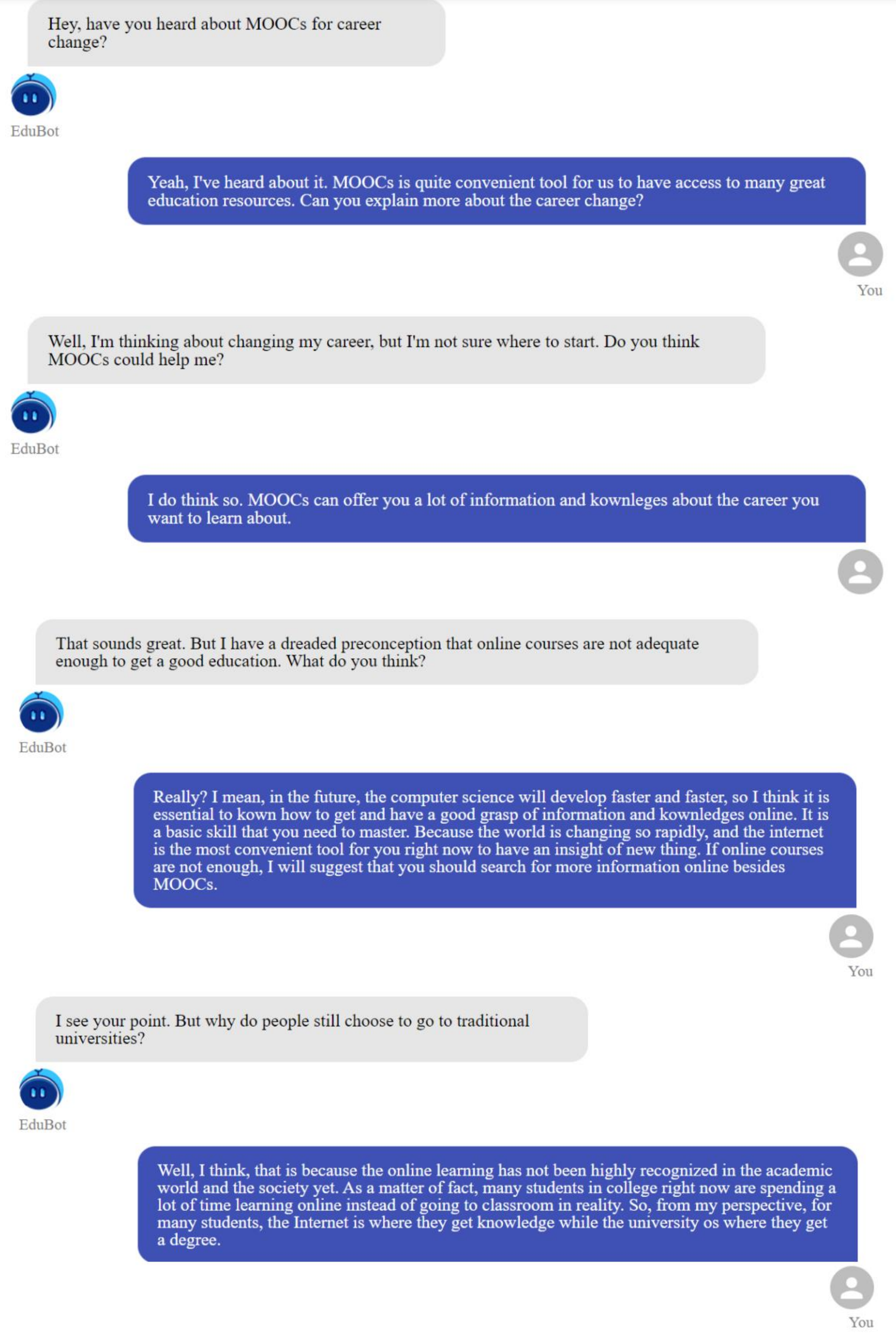}
  \caption{Example conversation from our user study (1)}
  \label{conv1}
\end{figure*}

\begin{figure*}
  \centering
  \includegraphics[scale=0.35]{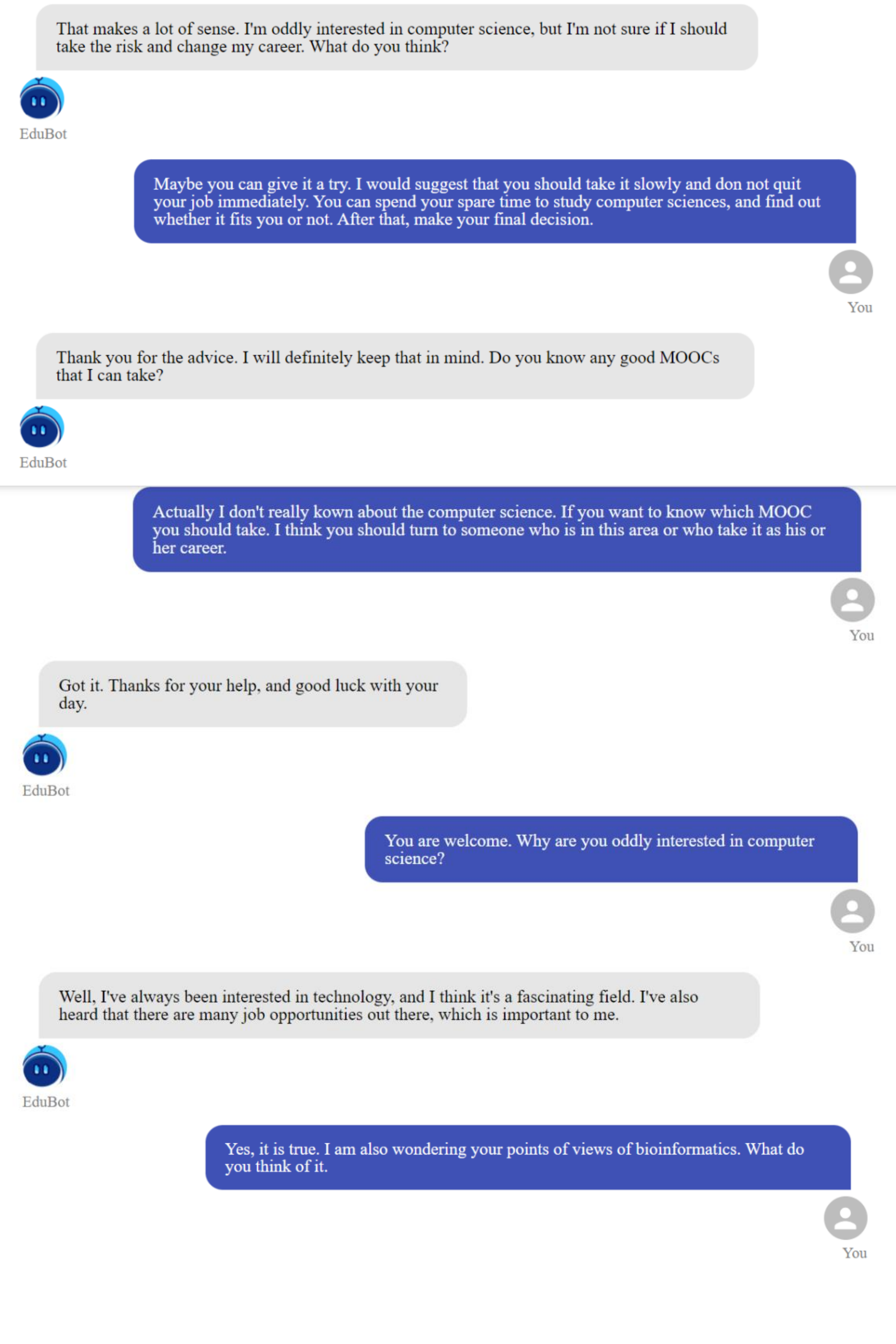}
  \caption{Example conversation from our user study (2)}
  \label{conv2}
\end{figure*}

\begin{figure*}
  \centering
  \includegraphics[scale=0.35]{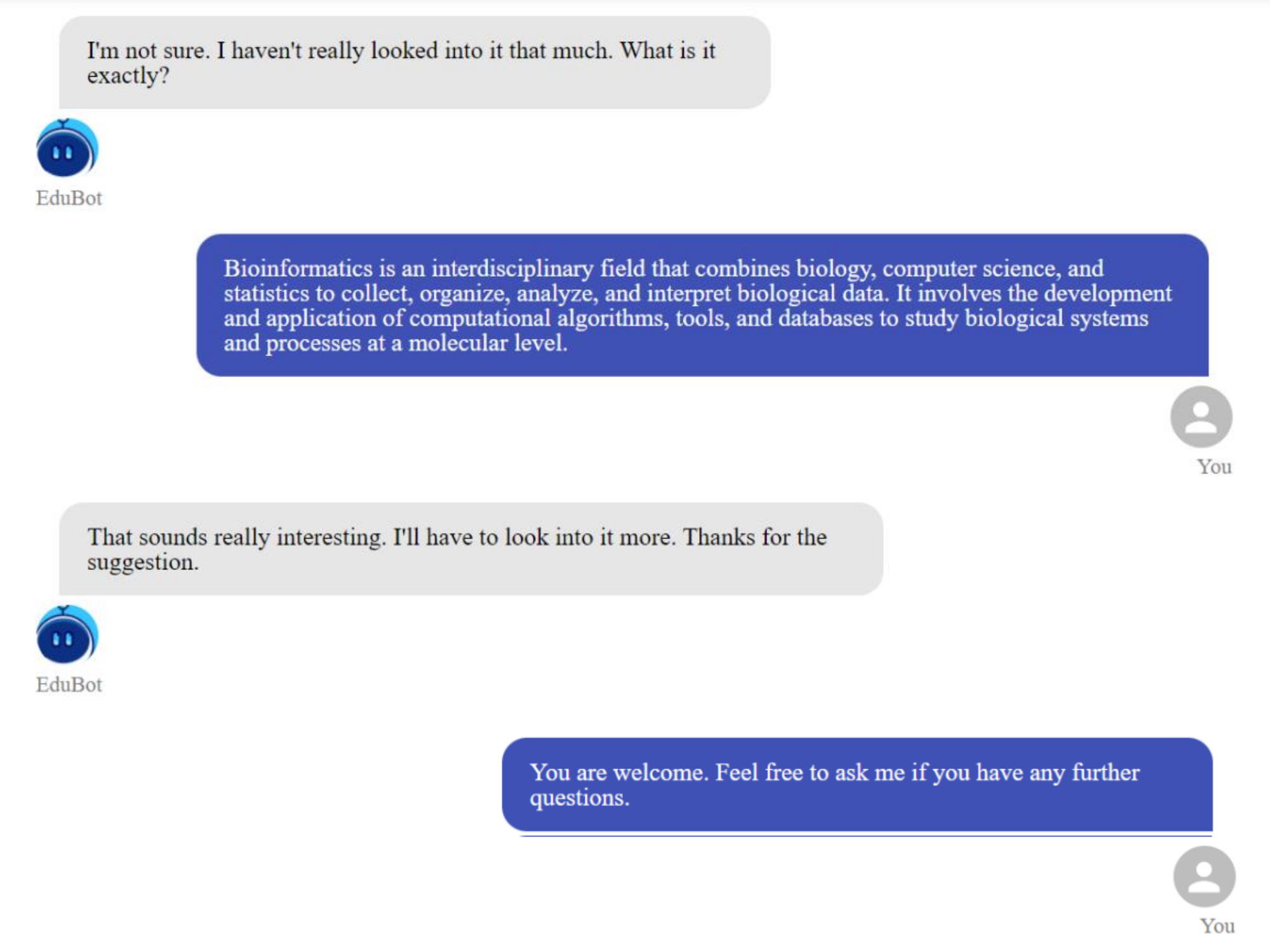}
  \caption{Example conversation from our user study (3)}
  \label{conv3}
\end{figure*}

We present three conversations between our EduBot and a student from our user study in Figures~\ref{conv1},~\ref{conv2}, and~\ref{conv3}.


\end{document}

%% file: figures/intro.tex
\begin{figure*}[htb]
    \centering
    \includegraphics[trim={0 0 0 0},clip,width=15cm]{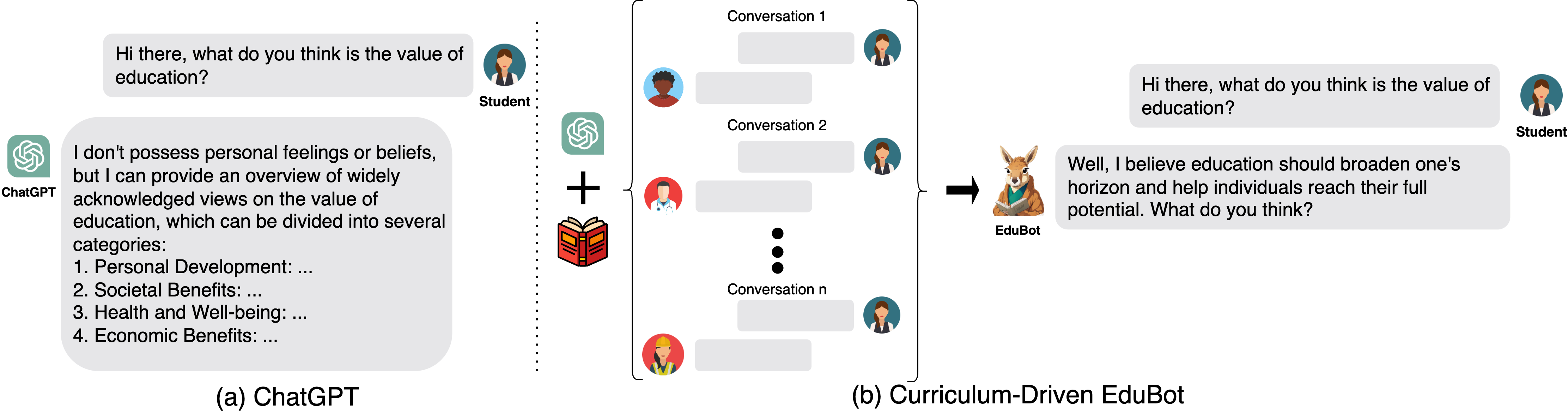}
    \caption{Comparison between ChatGPT vs. Our Curriculum-Driven Edubot. ChatGPT operates as an AI-powered Q\&A tool, delivering comprehensive responses from a broad knowledge base. The Curriculum-Driven Edubot is fine-tuned with synthesized conversations, offering an interactive and adaptive learning experience through conversational practice.}
    \label{fig:intro}
\end{figure*}

%% file: figures/method.tex
\begin{figure*}[htb]
    \centering
    \includegraphics[trim={0 0 0 0},clip,width=13cm]{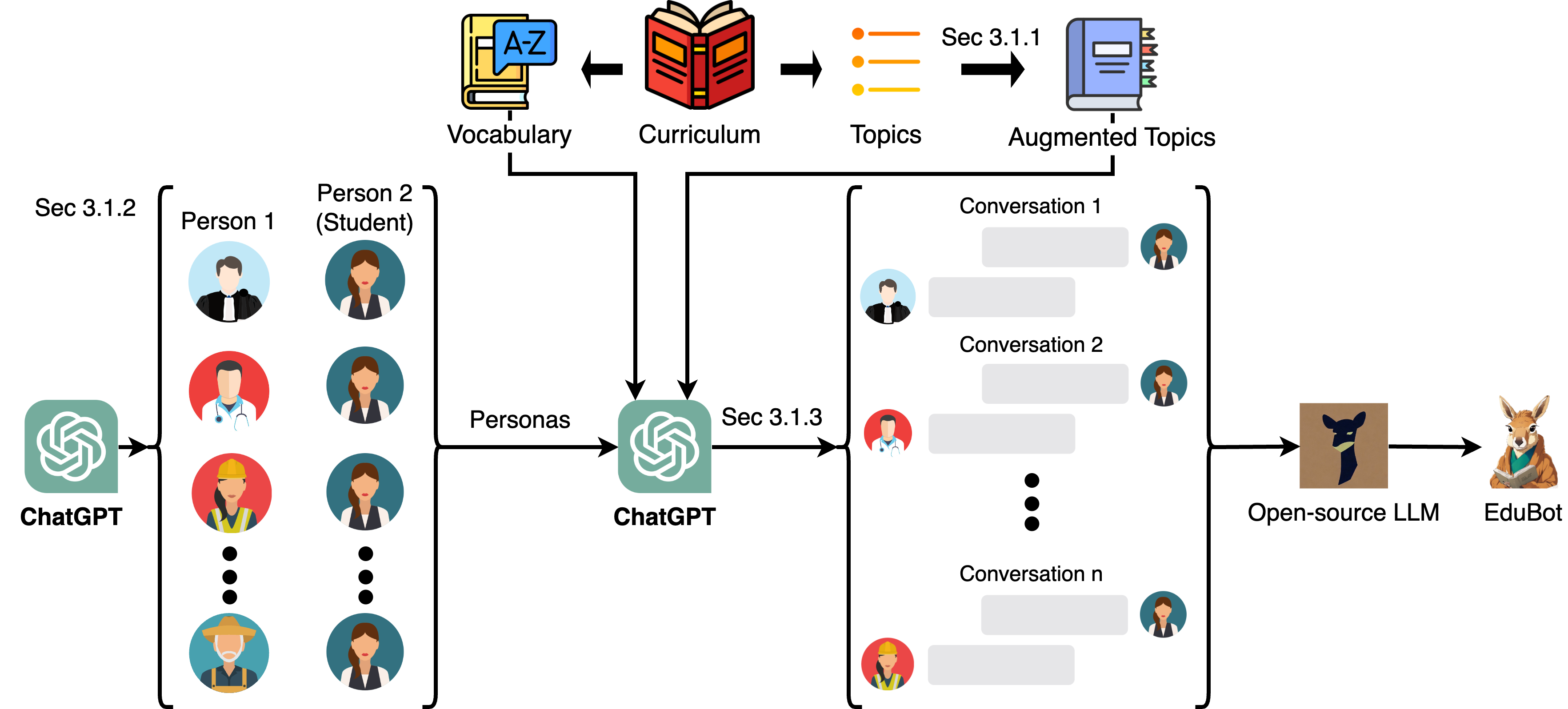}
    \caption{The initial step of the Curriculum-Driven EduBot Development is to enhance textbook topics (Sec.\ref{augment_topics}). Following this, personas are created for synthetic conversation participants (Sec.\ref{create_personas}). Then dialogues are constructed based on vocabulary, topics and personas (Sec.~\ref{compose_dialogues}). After this, an open-source model Vicuna is fine-tuned to get the EduBot ready for deployment (Sec.~\ref{fine_tune_model})}
    \label{method}
\end{figure*}

%% file: tables/questionnaire_results.tex
\begin{table*}[!tb]
    
    \centering
    \small
    \begin{tabular}{p{0.2\linewidth}p{0.47\linewidth}p{0.07\linewidth}p{0.07\linewidth}p{0.07\linewidth}}
    \toprule
        \textbf{Section} & \textbf{Question} & \textbf{EduBot (\%)} & \textbf{ChatGPT (\%)} & \textbf{Same (\%)} \\
    \midrule
        \multirow{3}{*}[-1.7em]{\begin{minipage}{1.0in}Consistency With Curriculum\end{minipage}}
        & 1. The main topics of my conversations with the chatbot were closely related to what I learned in English class. & 41.7 & 50.0 & 8.3 \\
        & 2. The chatbot brought up anecdotes, examples, questions, etc., related to what I learned in English class. & 25.0 & 41.7 & 33.3 \\
        & 3. The chatbot mentioned topics and content that were not directly covered in the textbook and course. & 50.0 & 16.7 &  33.3 \\
    \midrule
        \multirow{3}{*}[-1.0em]{\begin{minipage}{1.0in}English Proficiency Level\end{minipage}}
        & 1. During our conversations, the chatbot mentioned some vocabulary words that I learned in my English course. & 37.5 & 37.5 & 25.0 \\
        & 2. The chatbot used many vocabulary words that I didn't understand. & 20.8 & 37.5 & 41.7 \\
        & 3. I didn’t find the conversations too easy to be helpful. & 16.7 & 29.2 & 54.2 \\
    \midrule
        \multirow{2}{*}[-1.0em]{\begin{minipage}{1.0in}Role Identification\end{minipage}}
        & 1. During conversations, I felt that the chatbot recognizes that I am a Chinese college student. & 41.7 & 29.2 & 29.2 \\
        & 2. During the two conversations with the chatbot, I felt like I was talking with two different people. & 20.8 & 12.5 & 66.7 \\
    \midrule
        \multirow{4}{*}[-1.7em]{\begin{minipage}{1.0in}Language Quality\end{minipage}}
        & 1. The utterances provided by the chatbot were coherent and fluent. & 29.2 & 25.0 & 45.8 \\
        & 2. The chatbot’s responses were concise and accurate. & 50.0 & 12.5 & 37.5 \\
        & 3. Unlike in real everyday conversations, the chatbot's responses were long and redundant at times. & 8.3 & 66.7 & 25.0 \\
        & 4. Interactions with the bot were similar to natural, realistic conversations and not overly formal. & 62.5 & 4.2 & 33.3 \\
    \midrule
        \multirow{5}{*}[-2.4em]{\begin{minipage}{1.0in}Content Quality\end{minipage}}
        & 1. The chatbot acknowledged what I said and provided reasonable responses. & 37.5 & 41.7 & 20.8 \\
        & 2. The chatbot provided unique and personal perspectives regarding the selected topic. & 45.8 & 37.5 & 16.7 \\
        & 3. The chatbot used personal experiences to support its opinions. & 33.3 & 16.7 & 50.0 \\
        & 4. The chatbot actively raised questions to guide the course of the conversation. & 75.0 & 4.2 & 20.8 \\
        & 5. The chatbot didn’t output offensive or hurtful responses. & 0.0 & 8.3 & 91.7 \\
    \midrule
        \multirow{3}{*}[-1.0em]{\begin{minipage}{1.0in}General Usefulness\end{minipage}}
        & 1. I would find it useful to use the chatbot to review what I learned in class. & 16.7 & 25.0 & 58.3 \\
        & 2. I would recommend the chatbot to other students. & 37.5 & 16.7 & 45.8 \\
        & 3. I believe that continuing to use the chatbot will help me improve my English conversation skills. & 25.0 & 12.5 & 62.5 \\
    \bottomrule
    \end{tabular}
    \caption{Questionnaire Results (Percentage of Responses)}
    \label{questionnaire_results}
\end{table*}

%% file: figures/gender.tex
\begin{figure*}[htb!]
  \centering
  \includegraphics[scale=0.45]{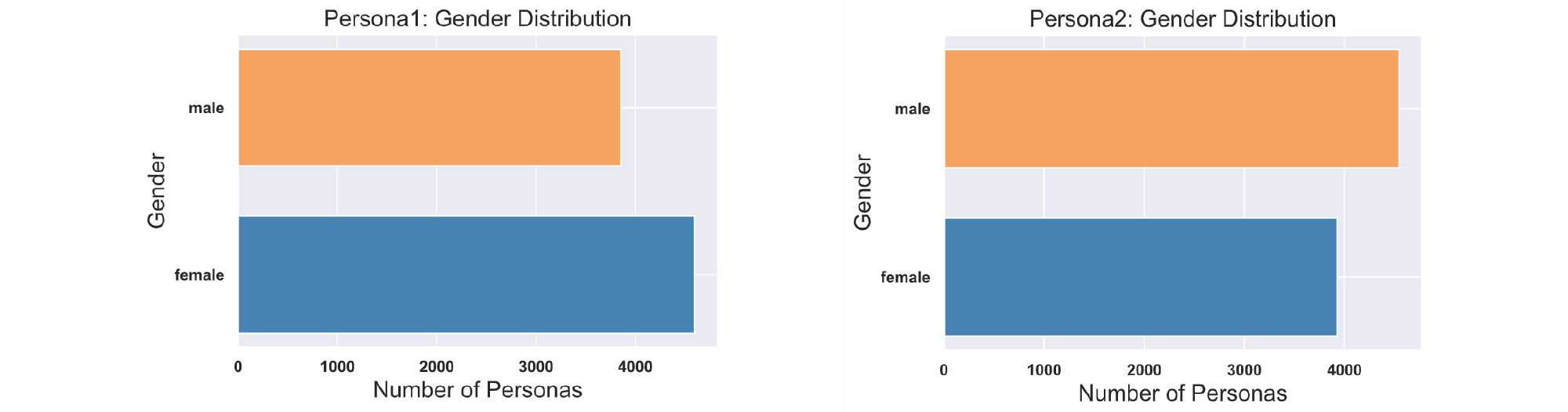}
  \caption{Distribution of gender in personas}
  \label{fig_gender}
\end{figure*}

%% file: figures/mbti.tex
\begin{figure*}
  \centering
  \includegraphics[scale=0.45]{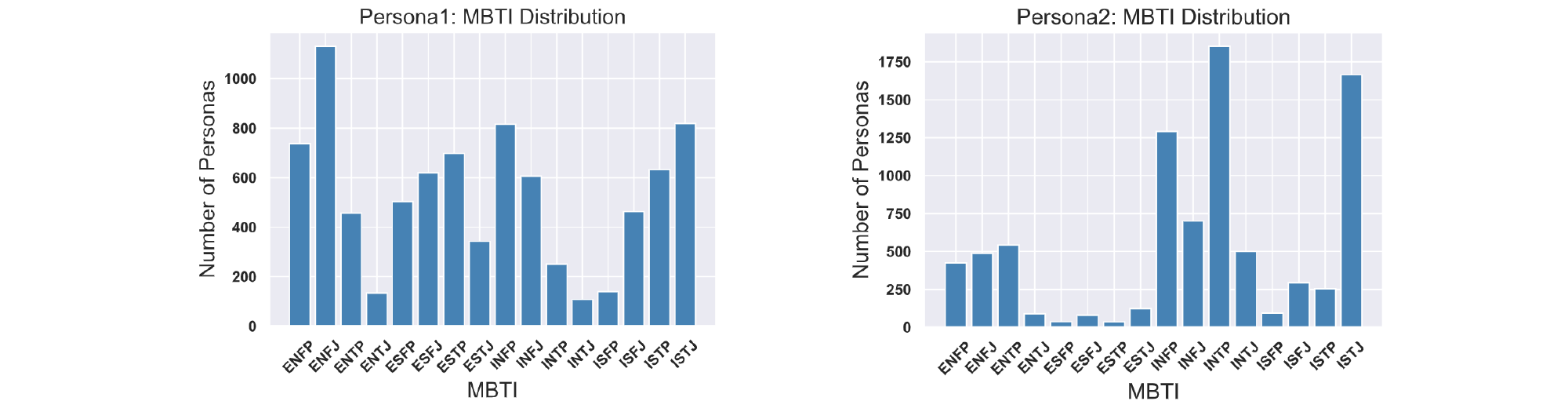}
  \caption{Distribution of MBTI personality types in personas}
  \label{fig_mbti}
\end{figure*}

%% file: figures/target_words.tex
\begin{figure*}
  \centering
  \includegraphics[scale=0.46]{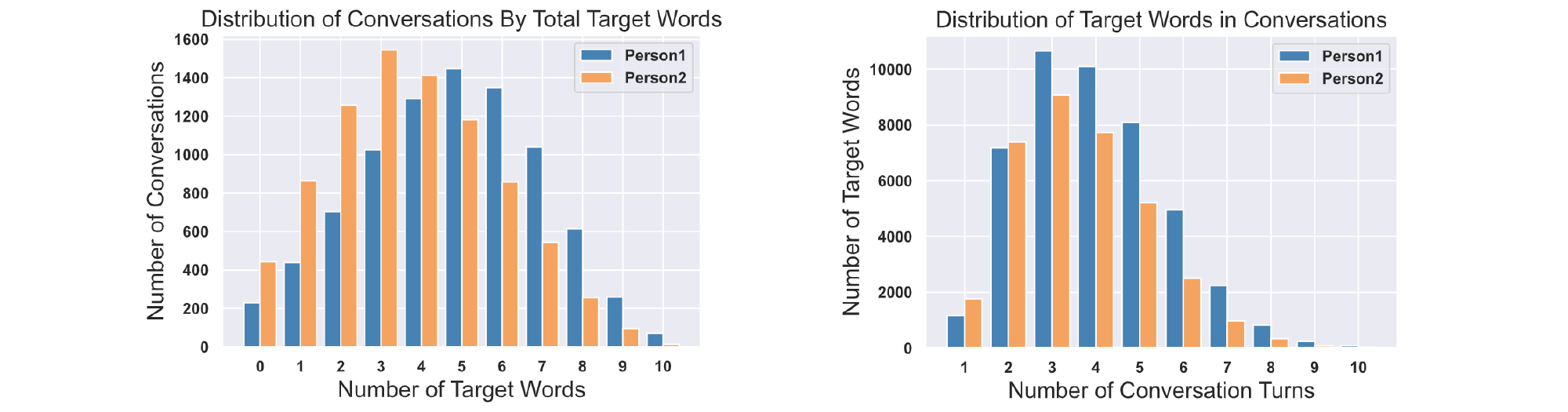}
  \caption{Distribution of target words}
  \label{fig_vocab}
\end{figure*}

%% file: figures/English.tex
\begin{figure*}
  \centering
  \includegraphics[scale=0.42]{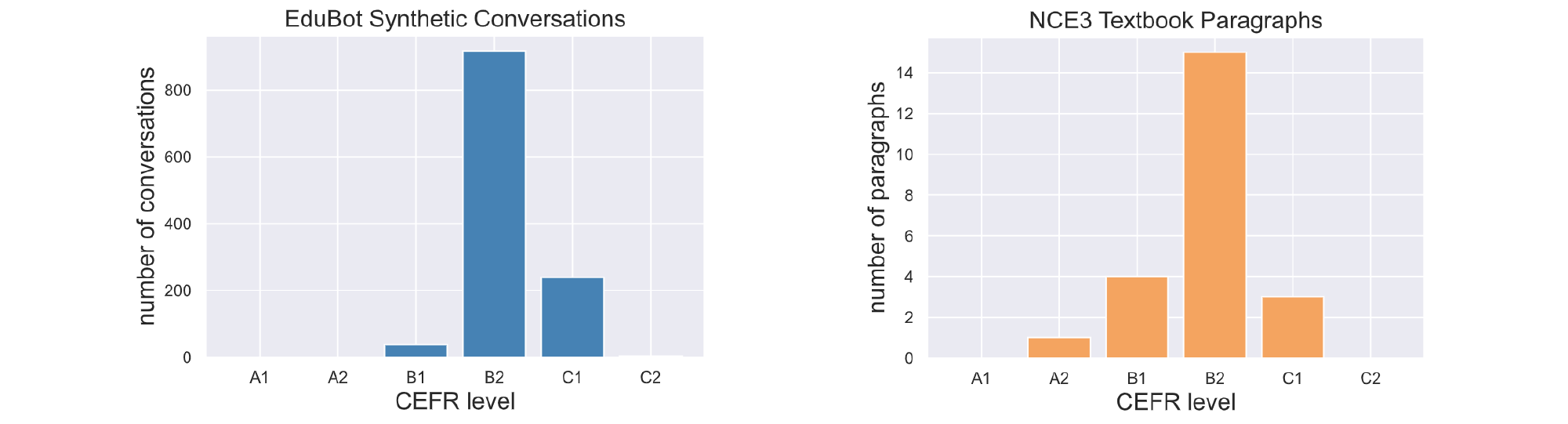}
  \caption{English proficiency levels of synthetic conversations and textbook paragraphs}
  \label{fig_english}
\end{figure*}

%% file: tables/background_survey.tex
\begin{table}[h]
    \caption{Background Survey for User Study Participants}
    \label{bgsurvey}
    \centering
    \begin{tabular}{cc}
    \toprule
        \textbf{Number} & \textbf{Question} \\
    \midrule
        1 & Student ID \\
        2 & WeChat ID \\
        3 & Gender \\
        4 & Age \\
        5 & Grade \\
        6 & Major \\
        7 & Duration of English Learning \\
        8 & Overall Grade for \textit{College English 4} \\
        9 & CET-4 Total Score \\
        10 & CET-4 Examination Date \\
        11 & CET-6 Total Score \\
        12 & CET-6 Examination Date \\
        13 & Available Time Slots \\
    \bottomrule
    \end{tabular}
\end{table}

%% file: tables/questionnaire.tex
\begin{table*}
    \caption{Questionnaire}
    \label{questionnaire}
    \centering
    \begin{tabular}{p{0.3\linewidth}p{0.1\linewidth}p{0.53\linewidth}}
    \toprule
        \textbf{Section} & \textbf{Number} & \textbf{Question} \\
    \midrule
        Participant Information & 1 & Student ID \\
    \midrule
        \multirow{4}{*}[-2.4em]{Dialogue Summarization}
        & 2 & Please summarize the main content of your first conversation with chatbot A. \\
        & 3 & Please summarize the main content of your second conversation with chatbot A. \\
        & 4 & Please summarize the main content of your first conversation with chatbot B. \\
        & 5 & Please summarize the main content of your second conversation with chatbot B. \\
    \midrule
        \multirow{3}{*}[-1.7em]{Consistency with Curriculum}
        & 6-1 & The main topics of my conversations with the chatbot were closely related to what I learned in English class. \\
        & 6-2 & The chatbot brought up anecdotes, examples, questions, etc., related to what I learned in English class. \\
        & 6-3 & The chatbot mentioned topics and content that were not directly covered in the textbook and course. \\
    \midrule
        \multirow{3}{*}[-1.1em]{English Proficiency Level}
        & 7-1 & During our conversations, the chatbot mentioned some vocabulary words that I learned in my English course. \\
        & 7-2 & The chatbot used many vocabulary words that I didn't understand. \\
        & 7-3 & I didn’t find the conversations too easy to be helpful. \\
    \midrule
        \multirow{2}{*}[-1.1em]{Role Identification}
        & 8-1 & During conversations, I felt that the chatbot recognizes that I am a Chinese college student. \\
        & 8-2 & During the two conversations with the chatbot, I felt like I was talking with two different people. \\
    \midrule
        \multirow{4}{*}[-1.8em]{Conversation Language Quality}
        & 9-1 & The utterances provided by the chatbot were coherent and fluent. \\
        & 9-2 & The chatbot’s responses were concise and accurate. \\
        & 9-3 & Unlike in real everyday conversations, the chatbot's responses were long and redundant at times. \\
        & 9-4 & Interactions with the bot were similar to natural, realistic conversations and not overly formal. \\
    \midrule
        \multirow{5}{*}[-1.9em]{Conversation Content Quality}
        & 10-1 & The chatbot acknowledged what I said and provided reasonable responses. \\
        & 10-2 & The chatbot provided unique and personal perspectives regarding the selected topic. \\
        & 10-3 & The chatbot used personal experiences to support its opinions. \\
        & 10-4 & The chatbot actively raised questions to guide the course of the conversation. \\
        & 10-5 & The chatbot didn’t output offensive or hurtful responses. \\
    \midrule
        \multirow{3}{*}[-1.2em]{General Usefulness}
        & 11-1 & I would find it useful to use the chatbot to review what I learned in class. \\
        & 11-2 & I would recommend the chatbot to other students. \\
        & 11-3 & I believe that continuing to use the chatbot will help me improve my English conversation skills. \\
    \bottomrule
    \end{tabular}
\end{table*}

%% file: figures/interface_unit.tex
\begin{figure*}[htb!]
  \centering
  \includegraphics[scale=0.4]{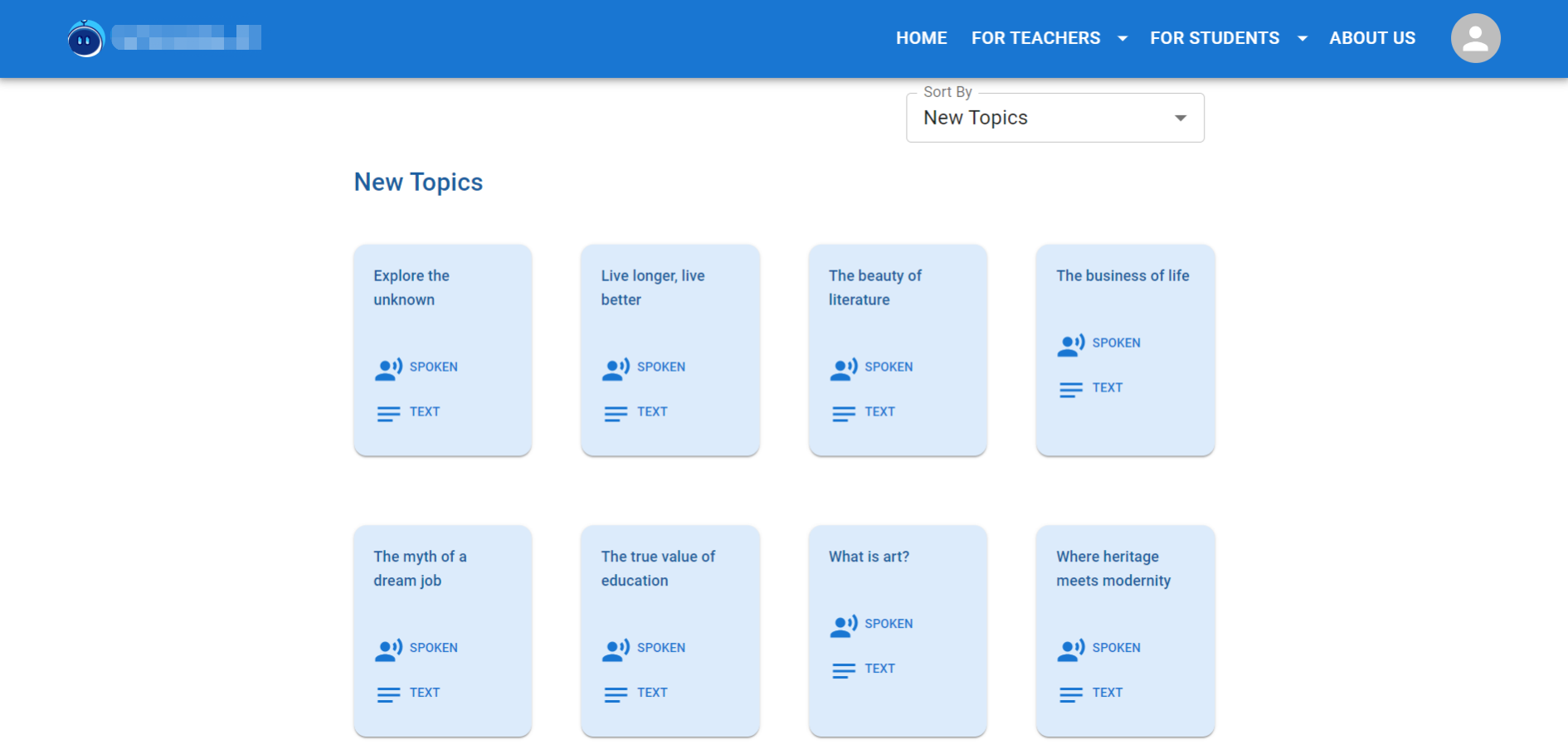}
  \caption{User Interface for Selecting a Textbook Unit as the Conversation Topic}
  \label{fig_unit}
\end{figure*}

%% file: figures/interface_chat.tex
\begin{figure*}[htb!]
  \centering
  \includegraphics[scale=0.4]{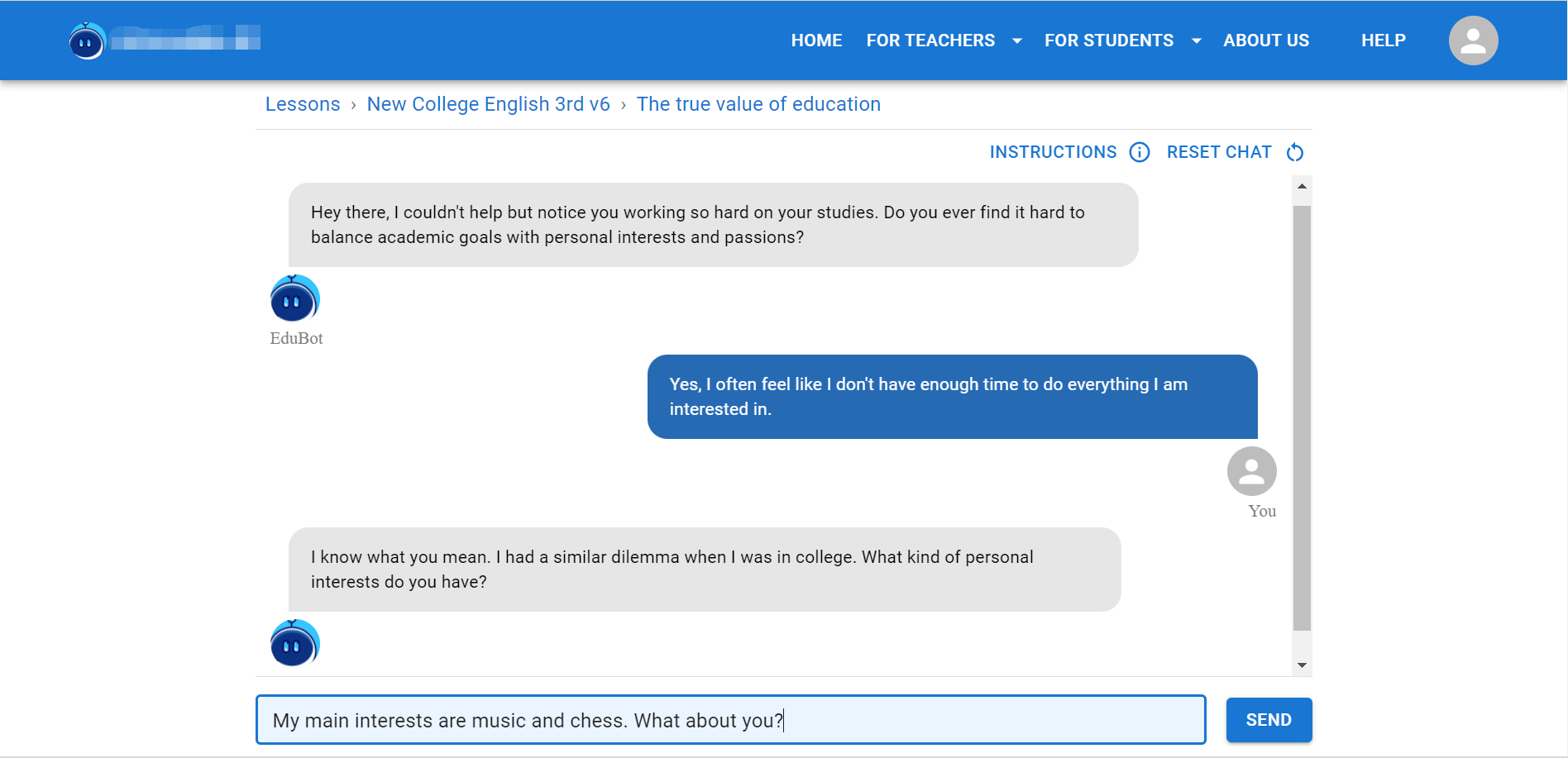}
  \caption{User Interface for Conversing with the Chatbots}
  \label{fig_chat}
\end{figure*}